\documentclass[letterpaper, 10 pt, conference]{ieeeconf}  

\IEEEoverridecommandlockouts                              
\usepackage{booktabs}
\usepackage{tabularx}
\usepackage{caption}
\usepackage{graphicx}
\usepackage{gensymb}
\usepackage{xcolor}
\usepackage{tikz}
\usepackage{verbatim}
\usepackage{hyperref}
\hypersetup{colorlinks,linkcolor={blue},citecolor={blue},urlcolor={black}}
\usepackage{blindtext}
\usepackage{subfigure}
\usepackage{array}

\usepackage{makecell, longtable}
\usepackage{stfloats}

\usepackage{amsmath}

\usepackage{color}

\newcommand{\tableLine}{\hline \\[\dimexpr-\normalbaselineskip+2pt]}
\newcommand{\tr}{\raggedright\arraybackslash}
\newcommand{\tc}{\centering\arraybackslash}

\title{\LARGE \bf
3D Face Recognition: A Survey
}

\author{Yaping~Jing,
        Xuequan~Lu,~\IEEEmembership{Member,~IEEE,}
        and~Shang~Gao
\thanks{Y. Jing, X. Lu and S. Gao are all with the School of Information Technology, Deakin University, Waurn Ponds, VIC, Australia. Corresponding author: X. Lu, S. Gao. \protect E-mails: \{jingyap, xuequan.lu, shang.gao\}@deakin.edu.au.}
}

\begin{document}

\maketitle
\thispagestyle{empty}
\pagestyle{empty}

\begin{abstract}

Face recognition is one of the most studied research topics in the community. In recent years, the research on face recognition has shifted to using 3D facial surfaces, as more discriminating features can be represented by the 3D geometric information. This survey focuses on reviewing the 3D face recognition techniques developed in the past ten years  which are generally categorized into conventional methods and deep learning methods. The categorized techniques are evaluated using detailed descriptions of the representative works. The advantages and disadvantages of the techniques are summarized in terms of accuracy, complexity and robustness to face variation (expression, pose and occlusions, etc). The main contribution of this survey is that it comprehensively covers both conventional methods and deep learning methods on 3D face recognition. In addition, a review of available 3D face databases is provided, along with the discussion of future research challenges and directions.

\end{abstract}



{\section{Introduction}\label{sec:introduction}}

Face recognition has become a commonly used biometric technology, which is widely applied in public records, authentication, security, intelligence and many other vigilance systems \cite{patil20153}. During the past decades, many 2D face recognition techniques have achieved high performance under controlled environments. The accuracy of 2D face recognition has been greatly enhanced especially after the emergence of deep learning. However, these techniques are still challenged by the intrinsic limitations of 2D images, such as illumination, pose, expression, occlusion, disguise, time delay and image quality \cite{zhou2014recent}. 3D face recognition may outperform 2D face recognition \cite{bowyer2006survey} with greater recognition accuracy and robustness, as it is less sensitive to pose, illumination, and expression \cite{huang2008labeled}. Furthermore, richer geometric information on 3D face can provide more discriminative features for face recognition. Thus, 3D face recognition has become an active research topic in recent years.

In 3D face recognition, 3D face models are normally used for training and testing purposes. Compared with 2D images, 3D face models contain more shape information. These rigid features can help face recognition systems overcome the inherent defects and drawbacks of 2D face recognition, for example, the facial expression, occlusion, and pose variations. Furthermore, a 3D model is relatively unchanged in terms of scaling, rotation, and illumination \cite{cai2019fast}. Most 3D scanners can acquire both 3D meshes/point clouds and corresponding textures. This allows us to integrate advanced 2D face recognition algorithms into 3D face recognition systems for better performance.

One of the main challenges of 3D face recognition is the acquisition of 3D images as it cannot be accomplished by crawling the Web like how 2D face images are collected. It requires special hardware equipment instead. According to the technologies used, it can be broadly divided into active acquisition and passive acquisition \cite{zhou20183d}. An active collection system actively emits invisible light (e.g. infrared laser beam) to illuminate the target face and obtain the shape features of the target by measuring the reflectivity. A passive acquisition system consists of several cameras placed separately from each other. It matches points observed from other cameras and calculates the exact 3D position of the matched point. The 3D surface is formed by a set of matched points. Since 2000, many researchers have begun to conduct an assessment of 3D face recognition algorithms on large-scale databases and published related 3D face databases, e.g. Face Recognition Vendor Tests (FRVT-2000) \cite{blackburn2001face}, FRVT-2002 \cite{phillips2003face}, the Face Recognition Grand Challenge (FRGC) \cite{phillips2005overview} and FRVT-2006 \cite{phillips2009frvt}. This suggests that there is a close relationship between large datasets and 3D face recognition techniques. In this paper, we also summarize the existing public 3D face databases and particularly their data augmentation methods when reviewing these recognition technologies.

There were relevant surveys conducted by researchers from different perspectives. In 2006, Bowyer et al. \cite{bowyer2006survey} reviewed the research trends in 3D face recognition. Abate et al. \cite{abate20072d} summarized the associated literature up to year 2007. Smeets et al. (2012) \cite{smeets2011comparative} studied various algorithms for expression invariant 3D face recognition and evaluated the complexity of existing 3D face databases. Followed by that, Zhou et al. \cite{zhou2014recent} categorized face recognition algorithms into single-modal and multi-modal ones in 2014. Patil et al. (2015) \cite{patil20153} studied the 3D face recognition techniques that comprehensively covered the conventional methods. Recently, \cite{soltanpour2017survey} and \cite{zhou20183d} both presented a review of the 3D face recognition algorithms, but only a few deep learning-based methods were covered. \cite{guo2019survey} and \cite{masi2018deep} reviewed the deep learning-based face recognition methods in 2018, but the focus was mainly on 2D face recognition. In this paper, we focus on 3D face recognition. Compared with the existing literature, the main contributions of our work are summarized as follows:

\begin{itemize}
\item To the best of our knowledge, this is the first survey paper that comprehensively covers conventional methods and deep learning-based methods for 3D face recognition.
\item Different from the existing surveys, it pays special attention to deep learning-based 3D face recognition methods.
\item It covers the latest and most advanced development in 3D face recognition, providing a clear progress chart for 3D face recognition.
\item It provides a comprehensive comparison of existing methods on the available datasets, and it suggests future research challenges and directions. 
\end{itemize}

\begin{figure}[ht]
    \centering
    \includegraphics[width=\linewidth]{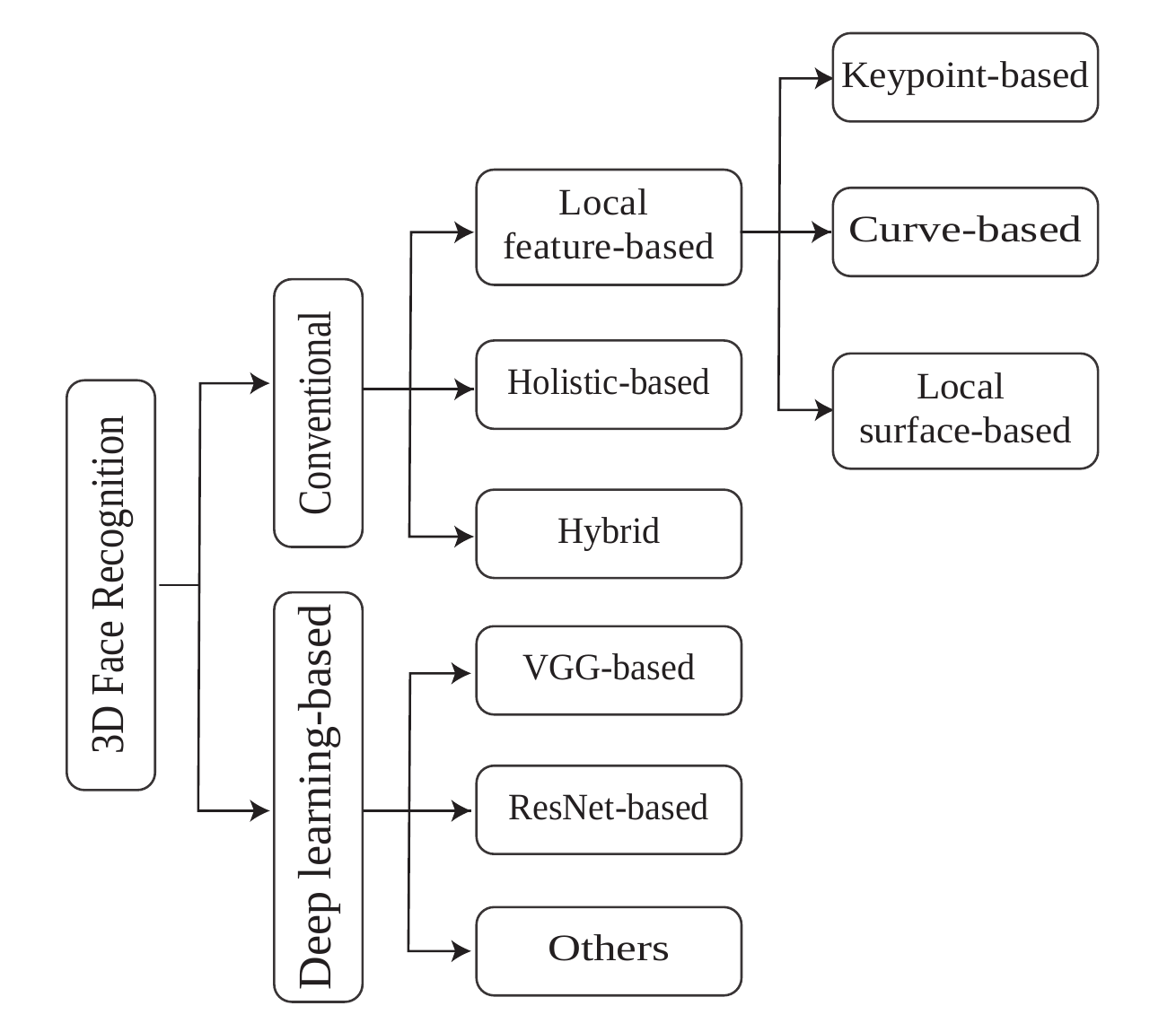}
   \caption{A taxonomy of 3D face recognition methods. }
   \label{fig:overview_taxonomy}
\end{figure}

According to the feature extraction methods adopted, 3D face recognition techniques can be divided into two categories: conventional method and deep learning-based method (Fig. \ref{fig:overview_taxonomy}). The conventional methods always use traditional algorithms to extract face features, e.g. Iterative closest point (ICP), principle component analysis (PCA), linear and nonlinear algorithms. They can be further divided into three types: local feature-based, holistic-based and hybrid. As for the deep learning-based methods, nearly all deep learning-based methods use pre-trained networks and then fine-tune these networks with the converted data (e.g. 2D images from 3D faces). Popular deep learning-based face recognition networks include VGGNet \cite{parkhi2015deep}, ResNet \cite{he2016deep}, ANN \cite{lawrence1997face} and recent lightweight CNNs such as MobileNetV2 \cite{howard2018inverted}. The structure of this paper is as follows. Section 2 introduces the widely used 3D face databases/datasets. Section 3 and 4 review conventional 3D face recognition methods and deep learning-based methods, respectively. Section 5 compares these methods and discuss future research directions, followed by a conclusion in Section 6.

\section{3D face database}
\label{sec:discussion}

\begin{figure}[ht]
\centering
\begin{minipage}[b]{0.3\linewidth}
{\includegraphics[width=1\linewidth]{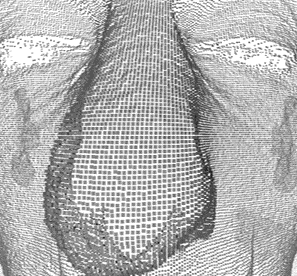}}
\end{minipage}
\begin{minipage}[b]{0.3\linewidth}
{\includegraphics[width=1\linewidth]{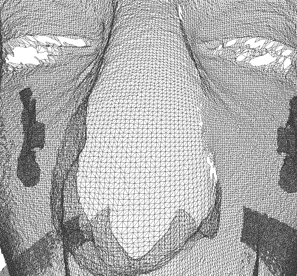}}
\end{minipage}
\begin{minipage}[b]{0.3\linewidth}
{\includegraphics[width=1\linewidth]{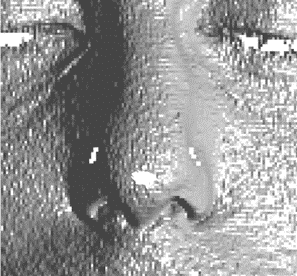}} 
\end{minipage} \\
\begin{minipage}[b]{0.3\linewidth}
\subfigure[Point cloud]
{\label{fig:data_representation_2a}\includegraphics[width=1\linewidth]{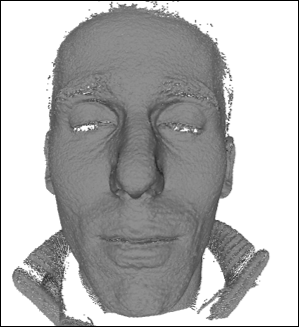}}
\end{minipage}
\begin{minipage}[b]{0.3\linewidth}
\subfigure[3D mesh]
{\label{fig:data_representation_2b}\includegraphics[width=1\linewidth]{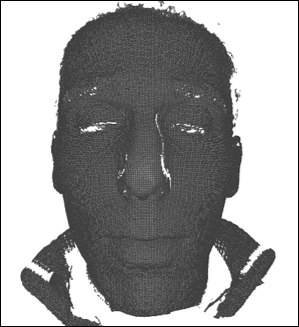}}
\end{minipage}
\begin{minipage}[b]{0.3\linewidth}
\subfigure[Range image]
{\label{fig:data_representation_2c}\includegraphics[width=1\linewidth]{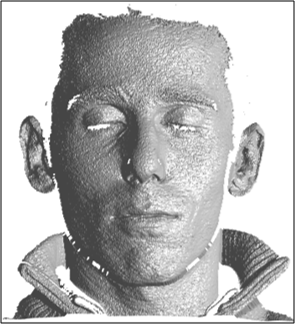}} 
\end{minipage}
\caption{3D face data representations \cite{colombo2011umb}. }
\label{fig:data_representation}
\end{figure}

Large-scale 3D face databases/datasets are essential for the development of 3D face recognition. They are used to train the feature extraction algorithms and evaluate their performance. To meet this demand, many research institutions and researchers have established various 3D face databases. Table \ref{table:database_tables} enlists the currently prominent 3D face databases and compares the data formats, the number of identities, image variations (e.g. expression, pose, and occlusion), and the scanner devices. There are four different 3D data formats: point cloud (Fig. \ref{fig:data_representation_2a}), meshes (Fig. \ref{fig:data_representation_2b}), range image (Fig. \ref{fig:data_representation_2c}) or depth maps, and 3D video; and two types of acquisition scanner devices: laser-based and stereo-based. Laser-based active collection systems use structured light scanners (e.g. Microsoft Kinect) or laser scanners (e.g. Minolta vivid scanner). Stereo-based devices are used by a passive acquisition system, such as Bumblebee XB3.

\begin{table*}
    \raggedright
    \caption{3D face databases. }\label{table:database_tables}
    \begin{tabularx}{\textwidth}{p{0.10\textwidth} p{0.03\textwidth} p{0.09\textwidth} p{0.03\textwidth} p{0.04\textwidth} p{0.06\textwidth} >{\tr}p{0.18\textwidth} >{\tr}p{0.1\textwidth} >{\tr}p{0.08\textwidth} >{\tr}p{0.08\textwidth}}
        \toprule
            Name/Reference & Year & Data type & IDs & Scans & Texture & Expression & Pose & Occlusion & Scanner \\
        \midrule
            3DRMA\cite{beumier2000automatic} & 2000 & Mesh & 120 & 720 & Yes & - & Slight left/right, up/down & - & Structured light \\ \tableLine
            
            FSU\cite{hesher2003novel} & 2003 & Mesh & 37 & 222 & No & - & - & - & Laser \\ \tableLine
            
            GavabDB\cite{moreno2004gavabdb} & 2004 & Mesh & 61 & 427 & No & Neutral, smile, accentuated laugh & $\pm30^{\circ}$ & - & Laser \\ \tableLine
            
            FRGC v2\cite{phillips2005overview} & 2005 & Range image & 466 & 4,007 & Yes & neutral, smiling & $\pm15^{\circ}$ & - & Laser \\ \tableLine
            
            UND\cite{chang2005evaluation} & 2005 & Range image & 275 & 670 & Yes & - & $\pm45^{\circ}$, $\pm60^{\circ}$ & - & Laser \\ \tableLine

            ZJU-3DFED\cite{wang2006exploring} & 2006 & Mesh & 40 & 360 & No & Neutral, smile, surprise, sad & - & - & Structured light \\ \tableLine
            
            BU3D-FE\cite{yin20063d} & 2006 & Mesh & 100 & 2,500 & Yes & Anger, happiness, sadness, surprise, disgust, fear & - & - & Stereo \\ \tableLine

            CASIA\cite{xu2006learning} & 2006 & Range image & 123 & 4,059 & No & Neutral, smile, eyes-closed, anger, laugh, surprise & $\pm90^{\circ}$ & - & Laser \\ \tableLine
            
            FRAV3D\cite{conde2006multimodal} & 2006 & Mesh & 105 & 1,696 & Yes & Neutral, smile, open mouth, gesture & Up and down, Y-axis turn, z-axis turn & - & Laser \\ \tableLine

            ND-2006\cite{faltemier2007using} & 2007 & Range image & 888 & 13,450 & Yes & Neutral, happiness, sadness, surprise, disgust, and other & $\pm15^{\circ}$ & - & Laser \\ \tableLine
            
            
            Bosphorus\cite{savran2008bosphorus} & 2008 & Point cloud & 105 & 4,666 & Yes & 34 & 3 yaw, pitch, cross rotations & 4 types & Stereo \\ \tableLine 
            
            UoY\cite{heseltine2008three} & 2008 & Mesh & 350 & 5,000 & Yes & Neutral, eyes closed, eyebrows raised, happy, anger & Frontal, up, down & - & Stereo \\ \tableLine
            
            SHREC08\cite{ter2008shape} & 2008 & Range image & 61 & 427 & No & Smile, laugh and arbitrary expressions & Front, up, down & - & - \\ \tableLine
            
            BJUT-3D\cite{baocai2009bjut} & 2009 & Mesh & 500 & 1,200 & Yes & - & - & - & Laser \\ \tableLine 
            
            Texas-3D\cite{gupta2010texas} & 2010 & Range image & 118 & 1,149 & Yes & Smiling, talking faces with open/closed mouths \& eyes & Frontal, $\pm10^{\circ}$ & - & Stereo \\ \tableLine
            
            UMBDB\cite{colombo2011umb} & 2011 & Range image & 143 & 1,473 & Yes & Neutral, smiling, angry, bored & Frontal & 7 types & Laser \\ \tableLine
        
            3D-TEC\cite{vijayan2011twins} & 2011 & Range image & 214 & 428 & Yes & Neutral, smiling & Frontal & - & Laser \\ \tableLine
            
            SHREC11\cite{veltkamp2011shrec} & 2011 & Range image & 130 & 780 & No & - & 5 directions & - & Laser \\ \tableLine
            
            NPU3D\cite{zhang2012npu} & 2012 & Mesh & 300 & 10,500 & No & 9 & 14 & 4 & Laser \\ \tableLine
           
            BU4D-FE\cite{zhang2013high} & 2013 & 3D video & 101 & 60,600 & Yes & - & - & - & Stereo \\ \tableLine
   
            KinectFaceDB\cite{min2014kinectfacedb} & 2014 & Range image & 52 & 936 & Yes & Neutral, smiling, mouth open & Left, right & Multiple & Kinect \\ \tableLine
            
            Lock3DFace\cite{zhang2016lock3dface} & 2016 & Range image & 509 & 5,711 & Yes & Happiness, anger, sadness, surprise, fear, disgust & $\pm90^{\circ}$ & Randomly cover-up & Kinect \\ \tableLine 
            
            F3D-FD\cite{urbanova2018introducing} & 2018 & Range image & 2,476 & - & Yes & - & Semi-lateral with ear & Half face & Stereo  \\ \tableLine 
                        
            LS3DFace\cite{zulqarnain2018learning} & 2018 & Point cloud & 1,853 & 31,860 & Yes & - & - & - & - \\ \tableLine

            WFFD\cite{jia20203d} & 2020 & Videos & 241 & 285 & Yes & - & - & - & - \\ \tableLine
            
            SIAT-3DFE\cite{ye2020siat} & 2020 & 3D & 500 & 8,000 & Yes & 16 & - & 2 & Structured light  \\ \tableLine
            
            FaceScape\cite{yang2020facescape} & 2020 & Videos & 938 & 18,760 & Yes & 20 & - & - & 68 DSLR cameras\\
        \bottomrule
     \end{tabularx}
\end{table*}

Before 2004, there were few public 3D face databases. Some representatives include \textbf{3DRMA} \cite{beumier2000automatic}, \textbf{FSU} \cite{hesher2003novel} and \textbf{GavabDB} \cite{moreno2004gavabdb}.The \textbf{GavabDB} database consists of 61 individuals, aged between 18 and 40. Each identity has 3 frontal images with different expressions and 4 rotating images without expressions \cite{moreno2004gavabdb}. In 2005, \textbf {FRGC V2.0} database was designed to improve the performance of face recognition algorithms, which had a huge impact on the development of 3D face recognition \cite {phillips2005overview}. So far, it is still used as a standard reference database (SRD) for evaluating the performance of 3D face recognition algorithms. In the same year, another important database \textbf{UND} (the University of Notre Dame) face database was released, where each identity has only one 3D image and multiple 2D images \cite{chang2005evaluation}.

From 2006 to 2010, there were more databases created. The largest one is \textbf{ND-2006} which is a superset of FRGC V2. It contains 13,450 images and 888 persons with as many as 63 images per identity \cite{faltemier2007using}. The second largest is \textbf{UoY} database, which consists of more than 5,000 models (350 people) owned by the University of York (UK) \cite{heseltine2008three}. The \textbf{CASIA} and \textbf{Bosphorus} database are similar in size, close to 5,000 images. The \textbf{CASIA} database was collected by using the non-contact 3D digitizer Minolta Vivid 910 in 2004 and contains 4,059 images of 123 objects \cite{xu2006learning}. It not only considers the individual variation of expressions, poses, and illumination but also introduces the combined changes of different expressions in different poses. \textbf{Bosphorus} has 381 individuals and the most expression and posture changes. It provides manual marking of 24 facial landmarks for each scanned image, such as nose tip, chin middle, eye corners \cite{savran2008bosphorus}. Another database which includes manual landmarks is \textbf{Texas-3D}. In Texas-3D, these 3D images have been preprocessed and 25 manually landmarks are provided. Therefore, It provides a good option for researchers to focus specifically on developing 3D face recognition algorithms, without considering the initial preprocessing of 3D images \cite{gupta2010texas}. The \textbf{BU3D-FE} (Binghamton University 3D Facial Expression) is a database specially developed for 3D facial expression recognition. It contains 100 identities with 6 expression types: anger, happiness, sadness, surprise, disgust and fear \cite{yin20063d}. For the \textbf{FRAV3D} database, 81 males and 24 females were involved, and three kinds of images (3D meshes, 2.5D range data, and 2D color images) were captured using the MINOLTA VIVID-700 red laser scanner \cite{conde2006multimodal}. \textbf{BJUT-3D} is one of the largest Chinese 3D face databases which includes 1,200 Chinese 3D face images \cite{baocai2009bjut}. The two smallest databases are \textbf{ZJU-3DFED} and \textbf{SHREC08}. The \textbf{ZJU-3DFED} database consists of 40 identities and 9 scans with four different kinds of expressions for each identity \cite{wang2006exploring}. The \textbf{SHREC08} database consists of 61 people with 7 scans for each \cite{ter2008shape}.

In the following five years after 2010, there were six remarkable databases created. The \textbf{UMBDB} database is an excellent database for testing the occlusion variance 3D face recognition algorithms, which contains 578 occlusion acquisitions \cite{colombo2011umb}. \textbf{3D-TEC} (3D Twins Expression Challenge) is a challenging dataset as it contains 107 pairs of twins with similar faces and different expressions \cite{vijayan2011twins}. Thus, this database is helpful to promote the performance of 3D face recognition technology. The \textbf{SHREC11} is based on a new collection of 130 masks with 6 3D face scans \cite{veltkamp2011shrec}. In addition to \textbf{BJUT-3D}, Northwestern Polytechnical University 3D (\textbf{NPU3D}) is another large-scale Chinese 3D face database, composed of 10,500 3D face data, corresponding to 300 individuals\cite{zhang2012npu}. The \textbf{BU4D-FE} is a 3D video database that records the spontaneous expressions of various young people by completing 8 emotional expression elicitation tasks \cite{zhang2013high}. The \textbf{KinectFaceDB} is the first publicly available face database based on the Kinect sensor and consists of four data modalities (2D, 2.5D, 3D, and video-based) \cite{min2014kinectfacedb}.

Recently, another large-scale 3D face database \textbf{Lock3DFace} was released. It is based on Kinect and contains various variations in expressions, poses, time-lapse, and occlusions \cite{zhang2016lock3dface}. \textbf{F3D-FD} is a large dataset which has the most individuals (2,476). For each individual, it includes partial 3D scans from the frontal and two semi-later views, and a one-piece face with lateral parts (including ears, earless, with landmarks) \cite{urbanova2018introducing}. The \textbf{LS3DFace} is the largest dataset so far, including 31,860 3D face scans of 1,853 identities. It is composed of multiple challenging public datasets, including FRGC v2, BU3D-FE, Bosphorus, GavabDB, Texas-3D, BU4D-FE, CASIA, UMBDB, 3D-TEC and ND-2006 \cite{zulqarnain2018learning}. The large-scale Wax Figure Face Database (\textbf{WFFD}) is designed to address the vulnerabilities in the existing 3D facial spoofing database and promote the research of 3D facial presentation attack detection \cite{jia20203d}. This database includes photo-based and video-based data. We only detail the video information in Table \ref{table:database_tables}. \textbf{SIAT-3DFE} is a 3D facial expression dataset in which every identity has 16 facial expressions including natural, happiness, sadness, surprise, several exaggerated expressions (open mouth, frown, etc.), and two occluded 3D models \cite{ye2020siat}. Another recent database is \textbf{FaceScape}, which consists of 18,760 textured 3D models with pore-level facial geometry \cite{yang2020facescape}.


It is well known that the performance of 3D face recognition algorithms could change on different 3D face databases. Increasing the gallery size could degrade the performance of face recognition \cite{gilani2016towards}. Although some algorithms have achieved good results on these existing 3D face databases, they still cannot be used in the real world due to more uncontrolled conditions in the real world. The establishment of large-scale 3D face databases to simulate real situations are very essential to facilitate the research of 3D face recognition. In addition, collecting 3D face data is a very time-consuming and source-demanding task. Research on large dataset generating algorithms would be one of the future works.

\section{Conventional methods}
\label{sec:conventional}

\begin{figure*}[ht]
    \centering
    \includegraphics[width=0.95\textwidth]{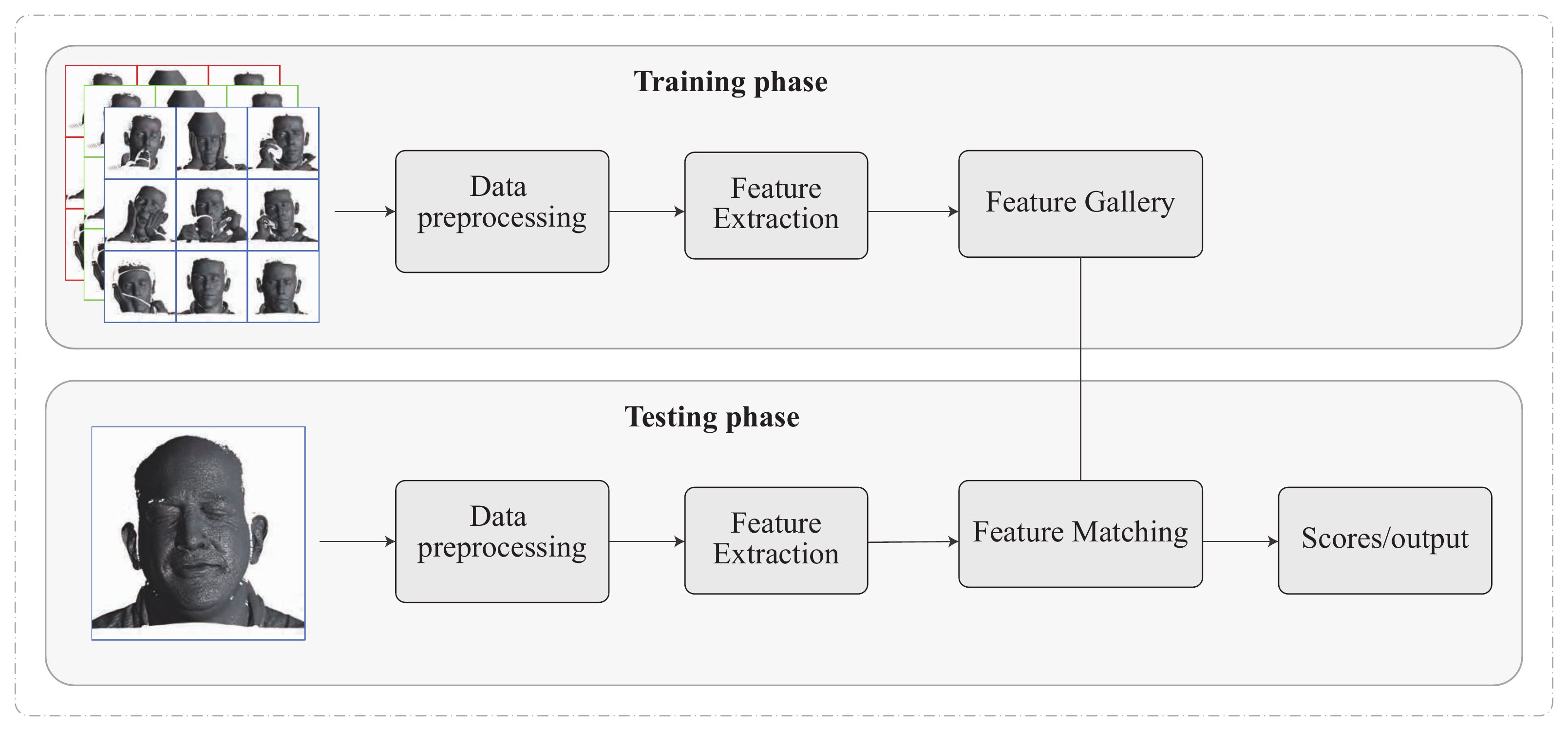}
    \caption{The pipeline of 3D face recognition. }
    \label{fig:conventional_flow}
\end{figure*}

In a conventional 3D face recognition system, there are two main phases: training and testing, as shown in Fig. \ref{fig:conventional_flow}. In the training phase, 3D face data is required to generate a feature gallery. Facial features are obtained through a data preprocessing and feature extraction model before being saved in the feature gallery. In the testing phase, a probe is acquired as the target face, and the same data preprocessing and feature extraction process as the training phase is performed. Face recognition is a matching process. The feature vectors of the target face are compared with the feature vectors stored in the feature gallery. The gallery is scanned and returns the face which has the closest matching distance. If the distance is lower than a predefined threshold, the target face is marked as recognized, otherwise, it fails. Thus, a face recognition process contains three core steps: data preprocessing, feature extraction and face matching. All of them can influence the performance of recognition.

\subsection{Data Preprocessing and matching}
In most situations, the acquired raw 3D face data cannot be directly used as the input for feature extraction systems as it may contain redundant information \cite{zhou20183d}, for example, hair, neck, and background context. This information will influence the accuracy of recognition. Thus, the 3D data is usually preprocessed before passing into a feature extraction model. In general, the data preprocessing could include three main parts: facial landmarks detection and orientation, data segmentation and face registration. Facial landmarks are a set of keypoints defined by anthropometric studies \cite{farkas1994anthropometry}, which can be used to automatically localize and register a face. Some databases already provide the landmarks of a face image. Data segmentation is the process of utilizing facial landmarks such as the nose tip and eye corners, to segment the facial surface \cite{farkas1994anthropometry}. This process is always used for local-based methods, which extract identifiable facial parts like the nose and eyes part for feature extraction. As an essential step before feature extraction and matching, face registration is to convert the target surface (entire face or face parts) to align with the training surface in the gallery.

After extracting the feature vectors from the original face, comes the most important part - face matching. The distances between the target face and the stored features in gallery are calculated. The common metrics include the Euclidean distance, Hausdorff distance, and angular Mahalanobis distance.


According to Zhao et al. \cite{zhao2003face} and our review on the last-decade literature, conventional face recognition algorithms can be classified into three types based on their feature extraction approaches: local feature-based, holistic-based, and hybrid, as shown in Fig. \ref{fig:overview_taxonomy}. Local-based approaches mainly focus on the local facial features such as nose and eyes \cite{zhao2003face}.In contrast to the local-based methods, the holistic-based approaches use the entire face to generate feature vectors for feature classification. Hybrid methods use both local and global facial features. 

For the local-based methods, fusion schemes are used to improve accuracy. There are five fusion schemes: sensor level fusion, feature level fusion, rank level fusion, decision level fusion, and score level fusion \cite{patil20153}. Sensor level fusion merges the original sensor data at the initial stage of recognition; feature level fusion involves the combination of features extracted from different facial representations of a single object; for rank level fusion, ranks are assigned to gallery images based on a descending sequence of confidence; score level fusion is a combination of the matching scores of each classifier based on a weighting scheme; decision-level fusion combines the decision of each classifier \cite{patil20153}.

Details of the three conventional face recognition methods are discussed below.

\subsection{Local feature-based methods}
\label{sec:local}
In the last decade, many local feature-based approaches were built, where local feature descriptors were used to describe the 3D local facial information. Table \ref{table:local_methods} enlists the remarkable 3D local-based methods and summarizes their important details. According to \cite{soltanpour2017survey}, these methods can be classified into three different types based on the descriptors: keypoint-based, curve-based and local surface-based. For the keypoint-based methods, a set of 3D keypoints are detected based on the face geometric information and used to build feature descriptors by calculating relationships between these keypoints; The curve-based methods use a set of curves on one face surface as features vectors; The local surface-based methods extract features from some regions of face surface \cite{soltanpour2017survey}.


\subsubsection*{Keypoint-based}
A keypoint-based method has two important steps: keypoint detection and feature descriptor construction \cite{soltanpour2017survey}. It uses a set of keypoints and their geometric relationships to represent facial features. Therefore, it can partially process a face image with missing parts or occlusions. However, due to using a large number of keypoints, it also involves higher computational cost. The most effective keypoint selection is also crucial for creating effective feature vectors.

One of the most commonly used keypoint detectors is Scale Invariant Feature Transformation (SIFT) \cite{lowe2004distinctive}. For example, \cite{berretti20113d} used SIFT to detect relevant keypoints of a 3D depth image, where local shape descriptors were adopted to measure the changes of face depth in the keypoints neighborhood. In \cite{inan20123}, SIFT descriptors were applied to 2D matrices including shape index, curvedness, Gaussian and mean curvature values generated from 3D face data to obtain feature vectors. \cite{soltanpour2016multimodal} used SIFT keypoint detection on pyramidal shape maps to obtain 3D geometric information and combine it with 2D keypoints. However, this SIFT method is sensitive to pose changes. \cite{guo2016ei3d} used a 3D point cloud registration algorithm combining with local features to achieve both pose and expression invariance. Later, a Keypoint-based Multiple Triangle Statistics (KMTS) method was proposed by \cite{lei2016two} to address partial facial data, pose and large facial expression variations. Recently, SIFT was also used to detect keypoints in \cite{deng2017efficient}, which used local covariance descriptors and Riemann kernel sparse coding to improve the accuracy of 3D face recognition. The accuracy was further improved in \cite{deng2020multi}.

In order to improve the robustness to large occlusions or poses, SIFT keypoint detection is directly used for 3D mesh data. The extension of SIFT for 3D mesh is called MeshSIFT \cite{smeets2013meshsift}. In \cite{smeets2013meshsift}, salient points on 3D face surface are first detected as extreme values in a scale space, then an orientation is assigned to these points. A feature vector is used to describe them by concatenating the histograms of slant angles and shape indices. Before this approach was applied, \cite{li2011expression} also used minimum and maximum curvatures within a 3D Gaussian scale space to detect salient points, and used the histograms of multiple order surface differential quantities to characterize the local facial surface. The descriptors of detected local regions were further used in 3D face local matching. \cite{li2015towards} also described an extension to this work, in which a fine-grained matching of 3D keypoint descriptors was proposed to enlarge intra-subject similarity and reduce inter-subject similarity. However, a large number of keypoints were detected by these methods. A meshDOG keypoint detector was proposed by Ballihi et al. (\cite{berretti2013matching}, \cite{berretti2014selecting}). They first used the meshDOG keypoint detector and local geometric histogram (GH) descriptor to extract features, then selected the most effective feature based on the analysis of the optimal scale, distribution and clustering of keypoints, and the features of the local descriptor. Recently, \cite{elaiwat2015curvelet} exploited a curvelet-based multimodal keypoint detector and local surface descriptor that can extracts both texture and 3D local features. It reduces the computation cost of keypoint detection and feature builder as the curvelet transform is based on FFT.

In addition, a set of facial landmarks are used for creating feature vectors in some methods and shape index is widely used to detect landmarks. In \cite{creusot2011automatic}, keypoints were extracted from a shape dictionary, which was learned on a set of 14 manually placed landmarks on human face. As an extension to \cite{creusot2011automatic}, \cite{creusot2013machine} used a dictionary of L learned local shapes to detect keypoints, and evaluated them through linear (LDA) and nonlinear (AdaBoost). \cite{zhang2011robust} detected the resolution invariant keypoints and scale-space extreme on shape index images based on scale-space analysis, and used six scale-invariant similarity measures to calculate the matching score. In \cite{vezzetti20143d}, an entire geometry-based 3D face recognition method was proposed and 17 landmarks were automatically extracted based on the facial geometrical characteristics, which was further extended in \cite{vezzetti20183d}.

\subsubsection*{Curve-based}
A curve-based method uses a set of curves to construct feature descriptors. It is difficult to define whether it is local feature-based or holistic feature-based, because these curves usually cover the entire face, also they capture geometric information from different face regions to represent the 3D face. The curves can be grouped into level curves and radial curves according to their distribution. Level curves are closed curves with different lengths and no intersection. Radial curves are open curves, usually starting from the nose tip. 

The level curves can be further divided into equal iso-depth and iso-geodesic curves \cite{soltanpour2017survey}. The iso-depth curves can be obtained by transposing a plane across the facial surface in one direction and were first introduced by Samir et al. \cite{samir2006three}. \cite{samir2009intrinsic} expanded this work and proposed iso-geodesic curves which are level curves of a surface distance function from the nose tip. However, both of them are sensitive to occlusions, missing parts or larger facial expressions. Thus, radial curves were introduced in \cite{drira2010pose} and extended in \cite{drira20133d}. These curves can better handle the occlusions and missing parts as it is uncommon to lose a full radial curve and at least some parts of a radial curve can be used. Also, they can be associated with different facial expressions as the radial curves pass through different facial regions.

In \cite{li2012efficient}, facial curves in the nose region of a target face were extracted to form a rejection classifier, which was used to quickly and effectively eliminate different faces in the gallery. Then the face was segmented into six facial regions. A facial deformation mapping was produced by using curves in these regions. Finally, the adaptive regions were selected to match the two identities. In \cite{ballihi2012boosting}, geometric curves from the level sets (circular curves) and streamlines (radial curves) through the Euclidean distance functions of a 3D face were combined for high-accuracy face recognition . A highly compact signature of a 3D face can be characterized by a small set of features selected by Adaboost algorithm \cite{freund1999short}, which is a well-known machine learning-based feature selection method. Using the selected curves for face recognition, time was reduced from 2.64 seconds to 0.68 seconds. It was proved that the feature selection method could effectively improve system performance. To select high discriminative feature vectors and improve computation efficiency, Angular Radial Signatures (ARSs) was proposed by lei et al. \cite{lei2014efficient}. It was described as a set of curves emitting from the nose tip (as the origin of the facial range images) at intervals of $\theta$ radians.
 
Another type of facial curves was introduced by Berretti et al. \cite{berretti2012sparse}. It utilized SIFT to detect keypoints of 3D depth images and connected the keypoints to form the facial curves. A 3D face could be represented by a set of facial curves built by matched keypoints. In \cite{al2015novel}, There were also some extended applications of facial curves. 3D curves were formed by intersecting three spheres with the 3D surface and used to compute Adjustable Integral Kernels (RAIKs) in \cite{al2015novel}. A sequence of RAIKs generated from the surface patch around keypoints can be represented by 2D images so that the certain characteristics of the represented 2D images can have a positive impact on matching accuracy, speed and robustness. \cite{emambakhsh2016nasal} introduced Nasal patches and curves. First, seven landmarks on the nasal region were detected. A set of planes was created using pairs of landmarks. A set of spherical patches and curves were yielded by the intersection of these planes with the nasal surface to create the feature descriptor. Then the feature vectors were taken by concatenating histograms of x, y, and z components of the surface normal vectors of Gabor-wavelet filtered depth maps and were filtered by the genetic algorithm to select more stable features against facial expressions. Compared with previous methods, this method has shown high-class separability. Recently, \cite{abbad20183d} presented a geometry and local shape descriptor based on the Wave Kernel Signature (WKS) \cite{aubry2011wave} to overcome the distortions caused by face expressions.

\subsubsection*{Local surface-based}
One of the representative local feature-based methods is Local Binary Pattern (LBP), introduced by Ojala et al. \cite{ojala2002multiresolution}. It was initially used for 2D images. The local geometric features extracted from some regions of a face surface can be robust to the face expression variations \cite{soltanpour2017survey}. LBP was used to represent the facial depth and normal information of each face region in \cite{tang20133d}, where a feature-based 3D face division pattern was proposed to reduce the influence of facial local distortion. Recently, \cite{shi2020research} used the LBP algorithm to extract features of a 3D depth image, and used the SVM algorithm to classify them. The feature extraction time of each depth map in Texas-3D was reduced to 0.1856 seconds while \cite{smeets2013meshsift} needs 23.54 seconds. Inspired by LBP, \cite{li2014expression} proposed the multi-scale and multi-component local normal patterns (MSMC-LNP) descriptor, which can describe normal facial information more compactly and differently. The Mesh-LBP method was used in \cite{werghi2016boosting}, where LBP descriptors were directly applied on the 3D face mesh surface, fusing both shape and texture information.

Another type of local feature-based methods is based on geometric features. \cite{lei2013efficient} proposed a low-level geometric feature approach, which extracts region-based histogram descriptors from a facial scan. The feature regions include nose and eyes-forehead, which are comparatively less affected by the deformation caused by facial expressions. In this paper, a Support Vector Machine (SVM) and the fusion of these descriptors at both features and score level were applied to improve the accuracy. In \cite{tabia2014covariance}, a covariance matrix of the feature was used as the descriptor for 3D shape analysis, not the feature itself. Compared with feature-based vectors, covariance-based descriptors can fuse and encode all types of features into a compact representation \cite{hariri20163d}. Their work was expanded in \cite{hariri20163d}.

There are other local feature-based methods. In \cite {elaiwat20133}, local surface descriptors were constructed around keypoints, which were defined by checking the Curvelet coefficient in each subband. Each keypoint is represented by multiple attributes, such as Curvelet position, direction, spatial position, scale, and size. A set of rotation-invariant local features can be obtained by rearranging the descriptors according to the orientation of the key points. The method in \cite {ming2015robust} used the regional boundary sphere descriptor (RBSR), which reduced the computational cost and improved the classification accuracy. \cite{soltanpour2017high} proposed a local derivative mode (LDP) descriptor based on local derivative changes. It can capture more detailed information than LBP. An extension to this work was described in \cite{soltanpour2019weighted}. Recently, Yu et al. \cite{yu2018sparse} recommended utilizing the ICP (Iterative closest point) with resampling and denoising (RDICP) method to register each face patch to achieve high registration accuracy. With rigid registration, all face patches can be used to recognize the face, significantly improving the accuracy as they are less sensitive to expression or occlusion.

\begin{table*}[hb]
    \centering
    \caption{Local feature-based techniques. }\label{table:local_methods}
    \begin{tabular}{
        >{\tr}p{0.15\textwidth}
        >{\tr}p{0.13\textwidth}
        >{\tr}p{0.14\textwidth}
        >{\tr}p{0.12\textwidth}
        >{\tr}p{0.12\textwidth}
        >{\tr}p{0.09\textwidth}
        >{\tc}p{0.11\textwidth}}
        \toprule
        Author/year & Category & Methods & Advantage & Limitation & Database & RR (Rank-1, \%) \\ 
        \midrule
        Berretti et al. (2011) \cite{berretti20113d} & SIFT keypoint & Covariance matrix, $X^2$ dist & Partial facial & Keypoints redundancy & FRGC v2 & 89.2 (Partial faces) \\ \tableLine

        Li et al. (2011) \cite{li2011expression} & Mesh-based keypoint & Histograms, cosine dist & Expression & Pose & Bosphorus & 94.1 \\ \tableLine
        
        Creusot et al. (2011) \cite{creusot2011automatic} & Landmark keypoint & Linear combination & Expression & Computationally expensive & FRGC v2 & - \\ \tableLine

        
        Zhang et al. (2011) \cite{zhang2011robust} & Landmarks & SVM-based fusion, six similarity measures & Simple preprocessing, noise, resolution & Occlusions & FRGC v2 & 96.2 \\ \tableLine
        
        
        Inan and Halici (2012) \cite{inan20123} & SIFT keypoint & Cosine dist & Neutral expression & Noise & FRGC v2 & 97.5 \\ \tableLine
        
        Berretti et al. (2012) \cite{berretti2012sparse} & Curve & Sparse & Missing parts & Large pose, expression & FRGC v2\newline GavabDB\newline UND & \hfil95.6\newline\hspace*{10px}97.13\newline 75 \\ \tableLine
        
        Li and Da (2012) \cite{li2012efficient} & Curve & PCA & Expression, hair occlusion & Exaggerated expressions & FRGC v2 & 97.80  \\ \tableLine
        
        Ballihi et al. (2012) \cite{ballihi2012boosting} & Curve & Euclidean dist, AdaBoost & Efficient, data storage & Occlusions & FRGC v2 & 98 \\ \tableLine

                
        Berretti et al. (2013) \cite{berretti2013matching} & Mesh-based keypoint & $X^2$ dist & Missing parts & Low accuracy & UND & 77.1 \\ \tableLine
        
        Smeets et al. (2013) \cite{smeets2013meshsift} & Mesh-based keypoint & Angles comparison & Expression, partial data & Noise & Bosphorus\newline FRGC v2 & \hfil93.7\newline 89.6  \\ \tableLine
               
        Creusot et al. (2013) \cite{creusot2013machine} & Mesh-based landmark keypoint & Linear (LDA), non-linear (AdaBoost) & Expression & Complexity, occlusions  & FRGC v2\newline Bosphorus & \hfil- \newline -  \\ \tableLine
        
        Tang et al. (2013) \cite{tang20133d} & Local surface (LBP-based) & LBP, Nearest-neighbor (NN) & Expression & Occlusion, missing data & FRGC v2 & 94.89 \\ \tableLine
        
        Lei et al. (2013) \cite{lei2013efficient} & Local surface (Geometric feature) & SVM & Expression & Occlusion & FRGC v2\newline BU-3DFE & \hfil95.6\newline 97.7 \\ \tableLine 
        
        Elaiwat et al. (2013) \cite{elaiwat20133} & Local surface & Curvelet transform & Illumination, expression & Occlusion & FRGC v2 & - \\ \tableLine      
        
        Drira et al. (2013) \cite{drira20133d} & Curve & Riemannian framework & Pose, missing data & Extreme expression, complexity & FRGC v2 & 97.7 \\ \tableLine
        
        
        Li et al. (2014) \cite{li2014expression} & Local surface (LBP-based) & ICP, Sparse-based & Expression, fast & Pose, occlusion & FRGC v2 & 96.3 \\ \tableLine 
                
        Berretti et al. (2014) \cite{berretti2014selecting} & Mesh-based keypoint & Classifier & Occlusions, missing parts & Noise, low-resolution image & Bosphorus & 94.5 \\ \tableLine
        
        Lei et al. (2014) \cite{lei2014efficient} & Curve & KPCA, SVM & Efficient, expression & Occlusion & FRGC v2\newline SHREC08 & \hfil- \newline -  \\ 
       
    \bottomrule
    \end{tabular}
\end{table*}


 \begin{table*}[ht]
    \ContinuedFloat
    \centering
    \caption{Local feature-based techniques (continued). }\label{table:local_methods1}
    \begin{tabular}{
          >{\tr}p{0.15\textwidth}
        >{\tr}p{0.13\textwidth}
        >{\tr}p{0.14\textwidth}
        >{\tr}p{0.12\textwidth}
        >{\tr}p{0.12\textwidth}
        >{\tr}p{0.09\textwidth}
        >{\tc}p{0.11\textwidth}}
        \toprule  
        Tabia et al. (2014) \cite{tabia2014covariance} & Local surface (Geometric features) & Riemannian metric & Expression & Occlusion & GavabDB & 94.91 \\ \tableLine
        
        Vezzetti et al (2014) \cite{vezzetti20143d} & Landmark keypoint & Euclidean distance & Expression, occlusion & Low accuracy & Bosphorus  & - \\ \tableLine
        
        
       
        Li et al. (2015) \cite{li2015towards} & Mesh-based keypoint & Gaussian filters, fine-grained matcher & Expression, occlusion, registration-free & Cost & Bosphorus & 96.56 \\ \tableLine
        
        Elaiwat et al. (2015) \cite{elaiwat2015curvelet} & Mesh-based keypoint & Curvelet transform, cosine dist & Illumination, expressions & Occlusion & FRGC v2 & 97.1 \\ \tableLine
        
        Al-Osaimi (2015) \cite{al2015novel} & Curve & Euclidean dist & Fast, expression & Occlusion & FRGC & 97.78 \\ \tableLine
        
      Ming (2015) \cite{ming2015robust} & Local surface & Regional, global regression & Large pose, efficient & Patches detection & FRGC v2\newline CASIA\newline BU-3DFE & \hfil-\newline\hspace*{13px}-\newline-\\ \tableLine 
                
      Guo et al. (2016) \cite{guo2016ei3d} & keypoint & Rotational Projection Statistics (RoPS), average dist & Occlusion, expression and pose & Cost & FRGC v2 & 97 \\ \tableLine
        
      Soltanpour and Wu (2016) \cite{soltanpour2016multimodal} & SIFT keypoint & Histogram matching & Expression & Pose & FRGC v2 & 96.9 \\ \tableLine
       
      Lei et al. (2016) \cite{lei2016two} & SIFT keypoint & Two-Phase Weighted & Missing parts, occlusions, data corruptions & Extreme pose, expression & FRGC v2 & 96.3\\ \tableLine
        
      Emambakhsh and Evans (2016) \cite{emambakhsh2016nasal} & Curve & Mahalanobis, cosine dist & Expression, single sample & Occlusion & FRGC v2 & 97.9  \\ \tableLine
      
      Werghi et al. (2016) \cite{werghi2016boosting} & Local surface (LBP-based) & Cosine, $X^2$ dist & Expression, missing data & pose & BU-3DFE\newline Bosphorus & \hfil - \newline -  \\ \tableLine

      Hariri et al. (2016) \cite{hariri20163d} & Local surface (Geometric features) & Geodesic dist & Expression, pose & Partial occlusions & FRGC v2 & 99.2 \\ \tableLine
      
      
    
        Soltanpour et al. (2017) \cite{soltanpour2017high} & Local surface (LDP) & ICP & Expression & Extreme pose, missing data & FRGC v2\newline Bosphorus & \hfil 98.1\newline 97.3\\ \tableLine
     
        Deng et al. (2017) \cite{deng2017efficient} & SIFT keypoints & Riemannian kernel sparse coding & Low-complex & Expression, occlusion & FRGC v2 & 97.3\\ \tableLine        
              
        
        Abbad et al. (2018) \cite{abbad20183d} & Curve & Angles comparison & Expression, time consumption & Occlusions and missing data & GavabDB & 99.18\\ \tableLine
    
        Soltanpour et al. (2019) \cite{soltanpour2019weighted} & Local surface (LDP) & ICP & Computational cost & Pose & FRGC v2 & 99.3 \\ \tableLine
        
        Shi et al. (2020) \cite{shi2020research} & Local surface (LBP-based) & LBP, SVM & Low consumption & Pose, occlusions & Texas-3D & 96.83 \\
    \bottomrule
    \end{tabular}
\end{table*}

\subsubsection*{Summary}
Most local feature-based methods can better handle facial expression and occlusion changes as they use salient points and rigid feature regions, such as nose and eyes, to recognize one face. The main objective of local feature-based methods is to extract distinctive compact features \cite{soltanpour2017survey}. Table \ref{table:local_methods} summarizes the local feature-based methods, which are further categorized into keypoint-based, curve-based and local surface-based, as recapped below:

\begin{itemize}
\item The keypoint-based methods are assorted into three groups: SIFT-based, mesh-based, and landmarks. There are two important points worth noting: the selection of effective keypoints and construction of feature descriptor. If the amount of keypoints is too excessive, the computational cost will increase. However, if the keypoints are too sparse, some key features will be lost and the recognition performance may be affected. In addition, the algorithms for measuring the neighborhood of keypoints are very important as the geometric relationships of keypoints are used to build feature vectors.

\item The curved-based methods are broadly classified into level curve-based and radial curve-based methods. Since the level curves are sensitive to the occlusions and missing parts, most curve-based methods use radial curves. Generally, a reference point is required in a curved-based method. The nose region is rigid and contains more distinctive shape features than other regions, so it is used as the reference point in most curve-based methods \cite{soltanpour2017survey}. Therefore, the detection of nose tip is a crucial step in these methods. Its incorrect position can affect the extraction of curves and the performance of the face recognition system.

\item The local surface-based methods are divided into LBP-based, geometric features-based and others. Some of them also need high accuracy of the nose tip detection as the nose tip is used for face segmentation. Most local surface-based methods are robust to facial expressions and postures as the feature vectors are extracted from rigid regions of a face surface.

\end{itemize}

\subsection{Holistic-based methods}
\label{sec:global}

Compared with the local-based methods, holistic-based methods extract features from the entire 3D face surface. They are very effective and can perform well under the complete, frontal, and expression-invariance 3D faces. Common techniques used by holistic-based methods include ICP, Eigenfaces (PCA) and Fisherfaces. 


Table \ref{table:holistic_methods} summarizes the remarkable endeavours that have been made in this area. An intrinsic coordinate system for 3D face registration was proposed by Spreeuwers \cite{spreeuwers2011fast}. This system is based on a vertical symmetry plane passing through the nose, nose tip, and nose orientation. A 3D point cloud surface is transformed into a face coordinate system and PCA-LDA is used to extract features from the range image obtained from the new transformation system. \cite{ocegueda2011ur3d} presented a method named UR3D-C, which used LDA to train dataset and compress the biometric signature to just 57 coefficients. It still shows a high discriminant under the compact feature vectors. Bounding sphere representation (BSR), introduced in \cite{ming2012robust}, was used to represent both the depth and 3D geometric shape information by projecting the preprocessed 3D point clouds on the bounding spheres. Shape-based Spherical Harmonic Features (SHF) was proposed in \cite{liu2012learning}, where SHFs were calculated based on the spherical depth map (SDM). The SHF can capture the gross shape and fine surface details of a 3D face through the energies contained in the spherical harmonics at different frequencies. \cite{taghizadegan20123d} used 2DPCA to extract features and employed Euclidean distance for matching. In \cite{mohammadzade2012iterative}, the authors proposed a computationally efficient and simple nose detection algorithm. It constructs a low-resolution wide-nose Eigenface space using a set of training nose regions. The pixel in an input scan is verified as a nose tip if the mean square error between a candidate feature vector and its projection on the Eigenface space is less than a predefined threshold.

\cite{ming2014rigid} introduced a Rigid-area Orthogonal Spectral Regression (ROSR) method, where the curvature information is used to segment facial rigid area and OSR is used to extract discriminant feature. In \cite{ratyal20153d}, a 3D point cloud is registered in the inherent coordinate system with the nose tip as the origin, and a two-layer ensemble classifier is used for face recognition. A local facial surface descriptor was proposed by \cite{tang20153d}. This descriptor is constructed based on three principal curvatures estimated by asymptotic cones. The asymptotic cone is an essential extension of the asymptotic direction to the mesh model. It allows the generation of three principal curvatures representing the geometric characteristics of each vertex. \cite{gilani2017deep} proposed a region-based 3D deformable model (R3DM), which is formed from the densely corresponding faces. Recently, Kernel-based PCA is used for 3D face recognition. Due to the nature of face exhibiting non-linear shapes, non-linear PCA was used in \cite{peter20193d} to extract 3D face features as it has a notable benefits to data representation in high-dimensional space.

\begin{table*}
    \centering
    \caption{Holistic-based techniques. }\label{table:holistic_methods}
    \begin{tabularx}{\textwidth}{>{\tr}p{0.19\textwidth} >{\tr}p{0.18\textwidth} >{\tr}p{0.15\textwidth} >{\tr}p{0.15\textwidth} >{\tr}p{0.1\textwidth} >{\tc}p{0.11\textwidth}}
        \toprule
        Author/year & Methods & Advantage & Limitation & Database & RR (Rank-1, \%) \\ 
        \midrule
        Spreeuwers (2011) \cite{spreeuwers2011fast} & PCA-LDA & Less registration time & Expression, occlusions & FRGC v2 & 99 \\ \tableLine

        
        Ocegued et al. (2011) \cite{ocegueda2011ur3d} & Ll norm, ICP, LDA, Simulated Annealing algorithm & Speed efficient & Expression, occlusions & FRGC v2 & 99.7 \\ \tableLine
        
        Ming and Qiuqi (2012) \cite{ming2012robust} & Robust group sparse regression model (RGSRM) & Expression, pose & Distorted images & FRGC v2 \newline CASIA & \hfil-\newline - \\ \tableLine
         
        Peijiang et al. (2012) \cite{liu2012learning} & - & Faster, cost-effective & Expression, occlusion & SHREC2007\newline  FRGC v2 \newline Bosphorus & \hfil97.86\newline\hspace*{10px}96.94\newline 95.63 \\ \tableLine
        
        Taghizadegan et al. (2012) \cite{taghizadegan20123d} & PCA, Euclidean distance & Expression & Occlusion & CASIA & 98 \\ \tableLine
        
        Mohammadzade and Hatzinakos (2012) \cite{mohammadzade2012iterative} & PCA & Computation, expression & Occlusion, pose & FRGC & - \\ \tableLine

        Ming (2014) \cite{ming2014rigid} & PCA, Spectral Regression, the orthogonal constraint & Expression, computational cost, storage space & Occlusions & FRGC v2 & 95.24 \\ \tableLine
        
        Ratyal et al. (2015) \cite{ratyal20153d} & PCA, Mahalanobis Cosine (MahCos) & Pose, expression & Occlusion, missing part & GavabDB\newline FRGC v2 & \hfil100\newline 98.93 \\ \tableLine
 
        Tang et al. (2015) \cite{tang20153d} & principal curvatures & Computational cost & Expression, occlusion & FRGC v2 & 93.16  \\ \tableLine
        
        Gilani et al. (2017) \cite{gilani2017deep} & PCA, use CNN for landmark detection & Occlusions & Faster, expressions, poses & Bosphorus & 98.1 \\ \tableLine
        
        Peter et al. (2019) \cite{peter20193d} & Kernel-based PCA & Higher accuracy rate & - & FRGC v2 & -  \\  
        \bottomrule
    \end{tabularx}
\end{table*}

Based on the discussion above, most holistic-based methods have faster speed and lower computational complexity, but they are not suitable for handling occluded faces or missing part faces. In addition, variations in pose and scale may affect the recognition performance of global features, because the holistic-based algorithms create discriminating features based on all the visible facial shape information. This requires accurate normalization for pose and scale. However, it is not easy to obtain accurate pose normalization under noisy or low-resolution 3D scanning.

\subsection{Hybrid methods}
\label{sec:hybird}

Hybrid 3D face recognition methods combine different types of approaches (local-based and holistic-based) and apply both local and global features for face matching. They can handle more face variances such as expression, pose, and occlusion via combining different feature extraction techniques. Recent hybrid methods are compared in Table \ref{table:hybrid_methods}.

\cite{passalis2011using} used an automatic landmark detector to estimate poses and detect occluded areas, and used facial symmetry to deal with missing data. \cite{huang20123} proposed a hybrid matching scheme using multiscale extended LBP and SIFT-based strategy. In \cite{alyuz2012robust}, the problem of external occlusions was addressed and a two-step registration framework was proposed. First, a non-occluded model is selected for each face with the occluded parts removed. Then a set of non-occluded distinct regions are used to compute the masked projection. This method relies on the accurate nose tip detection. The performance is adversely affected if the data has some occlusions covering the nose area. \cite{alyuz20133} extended this work in 2013. \cite{fadaifard2013multiscale} proposed a scale-space based representation for 3D shape matching which is stable against surface noise.

In \cite{bagchi2014robust}, Bagchi et al. used ICP to register a 3D range image, and PCA to restore the occluded region. This method is robust to the noise and occlusions. Later, they improved the registration method and proposed an across-pose method in \cite{bagchi20153d}. \cite{liang2017pose} also proposed a 3D face recognition method with pose-invariant and a coarse-to-fine approach to detect landmarks under large yaw variations. At the coarse search step, HK curvature analysis is used to detect candidate landmarks and subdivide them according to the classification strategy based on facial geometry. At the fine search step, the candidate landmarks are identified and marked by comparison with the face landmark model. 

\begin{table*}
    \centering
    \caption{Hybrid techniques. }\label{table:hybrid_methods}
    \begin{tabularx}{\textwidth}{>{\tr}p{0.19\textwidth} >{\tr}p{0.18\textwidth} >{\tr}p{0.15\textwidth} >{\tr}p{0.15\textwidth} >{\tr}p{0.1\textwidth} >{\tc}p{0.11\textwidth}}
        \toprule
        Author/year & Methods & Advantage & Limitation & Database & RR (Rank-1, \%) \\ 
        \midrule
        Passalis et al. (2011) \cite{passalis2011using} & PCA & Pose, occlusion, missing data & Expression, low accuracy & UND & - \\ \tableLine 
        
       
        Zhang et al. (2012) \cite{huang20123} & SIFT-based, extended LBP & Registration-free (frontal) & Large pose (alignment required) & FRGC v2 & 97.6 \\ \tableLine

        Alyuz et al. (2012) \cite{alyuz2012robust} & ICP, PCA, LDA & Occlusions & Expression & Bosphorus & 83.99  \\ \tableLine
      
        Fadaifard et al. (2013) \cite{fadaifard2013multiscale} & L1-norm & Noise, computational efficiency & Occlusions, expression & GavabDB & 86.89 \\ \tableLine  
        
        Alyuz et al. (2013) \cite{alyuz20133} & ICP, PCA, LDA & Occlusions, missing data & Expression & Bosphorus\newline UMBDB & \hfil{-\newline -} \\ \tableLine 
                
        Bagchi et al. (2014) \cite{bagchi2014robust} & ICP, PCA & Pose, occlusions & Pose & Bosphorus  & 91.3 \\ \tableLine 

        Bagchi et al. (2015) \cite{bagchi20153d} & ICP, KPCA & Pose & Expression & GavabDB\newline Bosphorus\newline FRAV3D & \hfil96.92\newline\hspace*{10px}96.25\newline 92.25 \\ \tableLine 
       
        Liang et al. (2017) \cite{liang2017pose} & HK classification & Pose & Expression & Bosphorus & 94.79 \\
             
        \bottomrule
    \end{tabularx}
\end{table*}

Hybrid face recognition systems use both local features and global features. Thus, their structures may be more complicated than those of the local-based or holistic-based methods. The hybrid approaches could achieve better recognition accuracy at higher computational cost. In addition, similar to holistic-based methods, its face registration is a very important step, especially for overcoming the pose-variance and occlusion-variance.

\section{Deep learning-based 3D face recognition}
\label{sec:deeplearning}

In the last decade, deep neural networks have become one of the most popular techniques for face recognition. Compared with the conventional ones, deep learning-based methods have great advantages over image processing \cite{lecun2015deep}. For conventional methods, the key step is to find robust feature points and descriptors based on geometric information of 3D face data \cite{kim2017deep}. Compared with the end-to-end deep learning models, these methods have good recognition performance, but involve relatively complex algorithmic operations to detect key features \cite{kim2017deep}. While for deep learning-based methods, robust face representations can be learned by training a deep deep neural networks on large datasets \cite{kim2017deep}, which can hugely improve the face recognition speed.

\begin{figure*}[ht]
    \centering
    \includegraphics[width=1\textwidth]{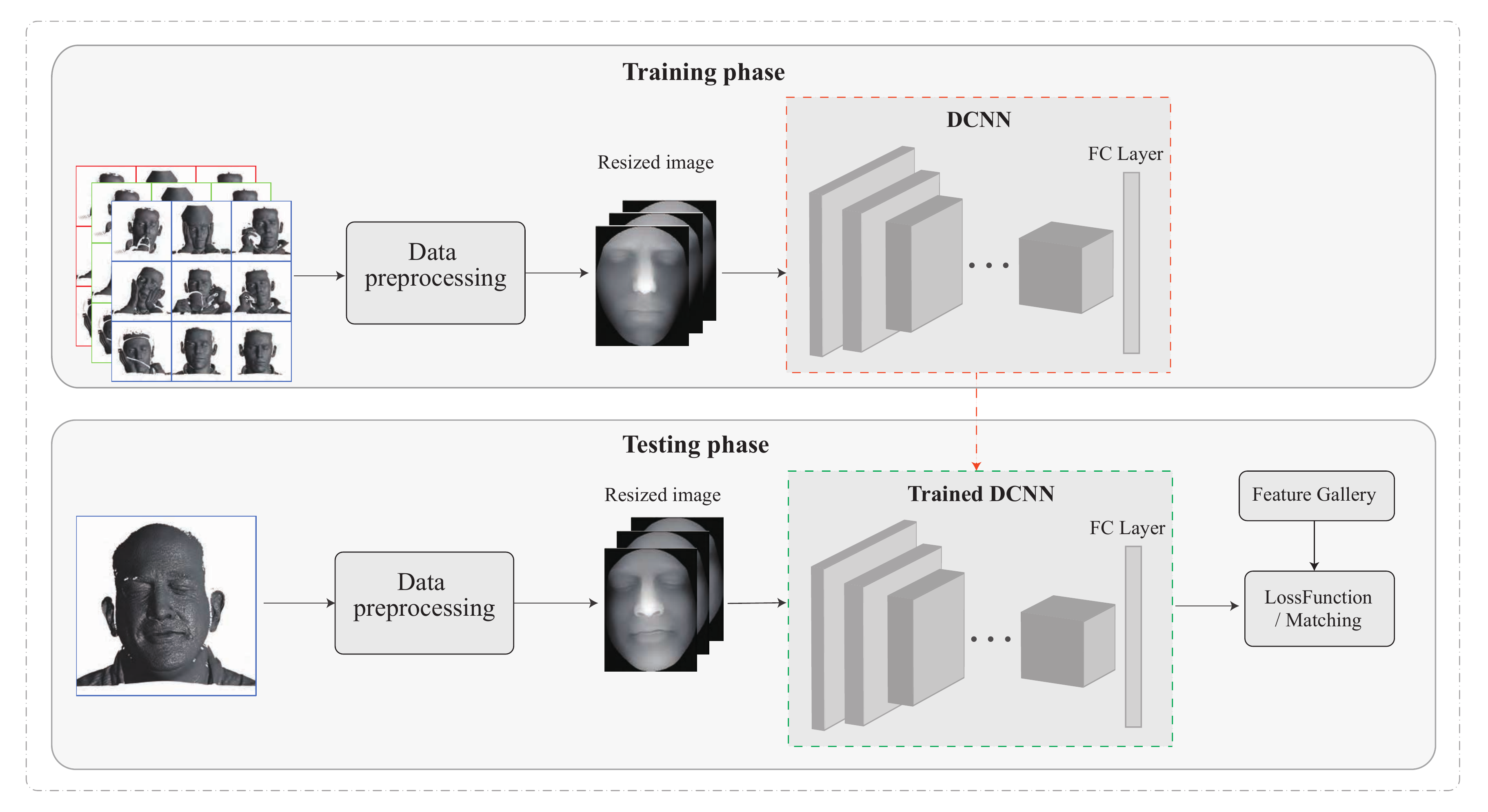}
    \caption{An overview of 3D deep learning-based face recognition methods. }
    \label{fig:deep_learning_flow}
\end{figure*}

There are a variety of deep neural networks for facial recognition and convolutional neural networks (CNN) are the most popular ones. A CNN usually consists of convolutional layers, pooling layers, and fully-connected (FC) layers. The purpose of a convolutional layer is to extract features from the input data. Each convolutional layer performs convolution operation with a filter kernel and applies a nonlinear transfer function. The objective of the pooling layers is to reduce the dimensions of the feature maps by integrating the outputs of neuron clusters of one layer into a single neuron in the next layer \cite{guo2019survey}. The robust and discriminative feature representation learned via CNN can significantly improve the performance of face recognition. Fig. \ref{fig:deep_learning_flow} depicts a common face recognition process based on Deep-CNN (DCNN). During the training phase, a set of training data is preprocessed first, such as alignment, resizing, etc., to generate a unified feature map and fit the input tensor of the DCNN architecture (e.g. the height, width, and channel of the feature map and the number of images). Then, the DCNN is trained by the preprocessed maps. In the testing phase, the feature representation of a probe is obtained from the trained DCNN and used to match with features in a given gallery. Before discussing the 3D face recognition methods using DCNN, we have a quick review on some typical 2D deep learning-based methods as their networks are still being used by some 3D methods.

\subsection{2D Face recognition} 
With the application of CNN, the performance of 2D face recognition systems (\cite{taigman2014deepface}, \cite{sun2014deep}, \cite{szegedy2015going}, \cite{parkhi2015deep}, \cite{schroff2015facenet}, \cite{he2016deep}) has been significantly improved. In these systems, face representations are directly learned from 2D facial images by training deep neural networks on large datasets. DeepFace \cite{taigman2014deepface} model is a nine-layer deep neural network which is trained on a labeled dataset including 4M facial images with over 4k identities. A 3D model-based alignment method is used and 97.35\% accuracy is achieved on the LFW \cite{huang2008labeled} dataset. The authors later extended this work in \cite{taigman2015web} and believed that the performance of CNN may reach a saturation point when the size of the training dataset increases.

The DeepId series methods (DeepID \cite{sun2014deep}, DeepID2 \cite{sun2014deep2}, DeepID2+ \cite{sun2015deeply}, DeepID3 \cite{sun2015deepid3}) extract deep features from various face regions. They incrementally improve the performance and reduce the error rate on the LFW dataset. \cite{szegedy2015going} proposed an inception DCNN architecture and created a 22-layers deep network named GoogLeNet model by repeating the Inception layers. The GoogLeNet model was trained on a large dataset with 200M face images of 8M identities. It utilizes metric learning algorithms and directly learns a mapping from face images to the compact Euclidean space. The embedding itself is optimized by a triplet loss during the network training. This is followed by VGG-Face \cite{parkhi2015deep}, which fine-tunes their model by a triplet-based metric learning method like FaceNet \cite{schroff2015facenet}. They also provided a large face dataset with 2.6M 2D images from 2,622 identities. Consequently, ResNet \cite{he2016deep} proposed a residual learning framework to simplify network training and evaluated the residual nets of up to 152 layers on the ImageNet dataset \cite{russakovsky2015imagenet}. Its depth is 8 times that of the VGG-Face \cite{parkhi2015deep}, but still shows lower complexity.

\subsection{3D Face recognition} 
As discussed above, deep learning-based 2D face recognition has made great achievements, and its performance is extremely high, almost close to 100\% on some specific databases (such as LFW). The high recognition rate of 2D face recognition proves that CNN-based methods are superior to the conventional feature extraction methods. Based on the intrinsic advantages of 3D faces relative to 2D faces in handling uncontrolled conditions such as pose, illumination and expression, more researchers are attracted to apply DCNN for 3D face recognition. Table \ref{table:deeplearning_methods} enlists the latest 3D face recognition techniques based on DCNN, and table \ref{table:deeplearning_performance} compares the recognition rate on different databases.

\subsubsection*{VGG-based}
\emph{Kim et al.} \cite{kim2017deep} proposed the first 3D face recognition model with DCNN. Their CNN architecture uses the VGG-Face \cite{parkhi2015deep} trained on 2D face images and then fine-tunes the CNN with augmented 2D depth maps. The last FC layer of the VGG-Face is replaced with a new last FC layer and a softmax layer. At the new last layer, weights are randomly initialized from a Gaussian distribution, with a mean of zero and a standard deviation of 0.01 \cite{kim2017deep}. In addition, the size of the dataset is expanded by augmenting the 3D point cloud of face scans with expression and pose variations during the training phase. In \cite{kim2017deep}, a multi-linear 3D Deformable Model (3DMM) is used to generate more expressions, including variations in both shape (${\alpha}$) and expression (${\beta}$). A 3D point cloud can then be represented by\cite{kim2017deep}:
\begin{equation}\label{eq:Kim}
\mathbf{X} = \mathbf{\overline{X}} + \mathbf{P}_s\alpha + \mathbf{P}_e\beta
\end{equation}
where $\mathbf{\overline{X}}$ is the average facial point cloud, $\mathbf{P}_s$ is the shape information provided by the Basel Face Model \cite{paysan20093d}, and $\mathbf{P}_e$ is the expression provided by FaceWarehouse \cite{cao2013facewarehouse}. The expression variations are created by randomly changing the values of expression ${\beta}$ parameters in the 3DMM. The randomly generated rigid transformation matrices are applied to an input 3D point cloud to demonstrate the pose variances \cite{kim2017deep}. At data preprocessing stage, a nose tip is first found in a 3D point cloud, then the 3D point cloud is cropped within a 100 mm radius. The rigid feature between the 3D face model and the reference face model is used to align the face 3D model. In order to fit the input size of its CNN architecture, the aligned 3D scan is orthogonally projected onto 2D image to generate a $224 \times 224 \times 3$ depth map. In addition, patches are randomly removed from the depth map to simulate hard occlusion. The model was evaluated on three public 3D databases: Bosphorus \cite{savran2008bosphorus}, BU3D-FE \cite{yin20063d} and 3D-TEC \cite{vijayan2011twins}, and the recognition rates were 99.2\%, 95.0\% and 94.8\%, respectively.


\emph{Deep 3D Face Recognition Network (FR3DNet)} \cite{zulqarnain2018learning} was trained on 3.1M 3D faces and specifically designed for 3D face recognition. It is also based on VGG face \cite{parkhi2015deep}. A rectifier layer is added for every convolutional layer. Compared with Kim's work \cite{kim2017deep}, a much larger dataset is generated and expanded by new identities. A new face $\mathbf{\hat{F}}$ is generated from a pair of faces ($\mathbf{F}_i$, $\mathbf{F}_j$) with the maximum non-rigid shape difference \cite{zulqarnain2018learning}: 
\begin{equation}\label{eq:zulqarnain}
\mathbf{\hat{F}} = \frac{\mathbf{F}_i + \mathbf{F}_j}{2} 
\end{equation}
These synthetic faces generated by this method have richer shape changes and details than the statistical face models \cite{paysan20093d}. However, the computational cost is very high as they are all generated from high dimensional raw 3D faces. In addition, 15 synthetic cameras were deployed at the front hemisphere of the 3D face to simulate pose variations and occlusions in each 3D scan. To fit the input of FR3DNet, 3D point cloud data is preprocessed to a $160 \times 160 \times 3$ image \cite{paysan20093d}. Before aligning and cropping the face, the point cloud is converted into a three-channel image. These three channels indicate three surfaces generated by using the gridfit (x, y grid) algorithm \cite{d2005surface}. They are the depth map z(x, y), azimuth map $\theta$(x, y) and elevation map $\phi$(x, y), where $\theta$ and $\phi$ are the azimuth and elevation angles of the normal vectors of 3D point cloud surface. The experiments were conducted on most public databases and the highest recognition accuracy was achieved on the BU3D-FE \cite{yin20063d} database, reaching 98.64\%.

\subsubsection*{ResNet-based}
\emph{Y. Cai, Y. Lei and M. Yang et al.} \cite{cai2019fast} designed three deep residual networks with different layers based on the ResNet \cite{he2016deep}, named as Pre-ResNet-14, Pre-ResNet-24 and Pre-ResNet-34. It is worth mentioning that a multi-scale triplet loss supervision is constructed by combining a softmax loss and the two triplet loss supervision on the last fully connected layer and the last feature layer \cite{cai2019fast}. To enlarge the size of the training set, the data was augmented in three ways: pose augmentation based on 3D scan, resolution and transformational augmentation based on range images \cite{cai2019fast}. For the preprocessing algorithm, raw 3D data is converted into a $96 \times 96$ range image and only the center of the two pupils and nose tip are used for alignment. For the preprocessing algorithm, three overlapping face components (the upper half face, the small upper half face, and only the nose tip) and one entire facial region are generated from the raw 3D data \cite{cai2019fast}. The most important part of this method is detecting the nose tip and two pupils. The three landmarks are detected from the 2D textured image of the corresponding 3D face data and are mapped to the 3D model. Then, a new nose tip is calculated by taking the highest point of the nose region 9 (centered on the tip of the nose with a radius of 25 mm). The nose tip is re-detected on the 3D model as the 2D domain detection could reduce detection accuracy due to pose variations. Another reason for detecting the nose tip by this means is that the lower dimensional feature vectors generated can be used to detect the new nose tip so that the computational cost can be reduced. Finally, the feature vectors of the four patches can be used alone or in combination for matching. It obtained high accuracy on four public 3D face databases: the FRGC v2, Bosphorus, BU-3DFE, and 3D-TEC datasets with 100\%, 99.75\%, 99.88\%, and 99.07\%, respectively.

\emph{Lin at al.} \cite{lin2019local} also adopted the ResNet-18 as their network. The big difference to other work is their data augmentation method. Instead of reconstructing 3D face samples from other raw data, they generated the the feature tensors directly based on the Voronoi diagram subdivision. The salient points are detected from a 3D face point cloud with its corresponding 2D face image and divided into 13 subdivisions based on the Voronoi diagram. The face can be expressed as $\mathbf{F} = [f_i, ... f_{13}]$ and the sub feature is $\mathbf{SubF}_i$. The feature tensor is extracted from a 3D mesh by detecting the salient points and integrating features of all the salient points, which can be represented as \cite{lin2019local}:
\begin{equation}\label{eq:lin2019}
\mathbf{F^k} = U_{i=1}^{13}SubF_i^k, k = 1,..,K
\end{equation}
where K is the number of 3D face samples of the same person. A new feature set could be synthesized by randomly choosing the $i^{th}$ sub feature set from the K samples. The network achieved very competitive performance on both Bosphrous and BU3D-FE databases with accuracy of 99.71\% and 96.2\% , respectively.


\emph{Tan at al.} \cite{tan2019face} designed a framework to specifically process the low-quality 3D data being captured by portable 3D acquisition techniques like mobile phones. The framework includes two parts: face registration and face recognition. At the face registration stage, a PointNet-like Deep Registration Network (DRNet) is used to reconstruct the dense 3D point cloud from low-quality sequences. The DRNet is based on ResNet-18 and takes a pair of $256 \times 256 \times 3$ coordinate-maps as input. To obtain the desired sparse samples from the raw datasets, noises and a random pose variation are added to the face scan. Then the new point cloud is projected onto a 2D plane with 1000 grids of the same size and a sparse face of 1,000 points is obtained by randomly selecting a point from each grid. Six sparse faces are generated from each face scan and passed to the DRNet to generate a new dense point cloud. Then the fused data is used as the input to Face Recognition Network (FRNet) which is also based on ResNet-18. Compared with FR3DNet, its facial recognition rate on UMBDB is higher, reaching 99.2\%.

\subsubsection*{Others}
\emph{Feng et al} \cite{feng20193d} adopted two DCNNs for feature extraction: one for color image, and the other for depth map built from 3D raw data. The output of the two feature layers was fused as the final input to an artificial neural network (ANN) recognition system. This ANN recognition system was tested on CASIA (V1) to compare the recognition rates by separately using the 2D feature layer, 3D feature layer, and the fusion of 2D and 3D features layers. A higher RR (98.44\%) was obtained with the fusion features.

\emph{Olivetti at al.} \cite{olivetti2019deep} proposed a method based on MobileNetV2. MobileNet is a comparatively new neural network specifically designed for mobile phones. It is easy to train and requires a low amount of parameters to tune. All their work was based on the Bosphorus database, which only contains 105 identities with 4,666 images. In order to obtain enough training samples, they augmented the data by rotating the original depth map (clockwise 25, counterclockwise 40) and creating a horizontal mirror for each depth map. The most important part of their work is the input data for DCNN. Geometric descriptors are used as input data instead of pure facial depth maps. The selection of geometric feature descriptors is based on the GH-EXIN network. The reliability of geometric descriptors based on curvature is proven in \cite{ciravegna2020assessing}. The input data is a three-channel image including the 3D facial depth map, the shape index and the curvedness, which can enhance the accuracy of the network. A 97.56\% recognition rate was achieved on the Bosphorus database.

\emph{Xu at al.} \cite{xu20193d} designed a dual neural network to reduce the impact of the number of training samples. The network consists of a dual-channel input layer that can fuse the 2D texture image and 3D depth map into one channel, and two parallel LeNet5-based CNNs. Each CNN processes the fused image separately to obtain its feature maps, which are used to calculate the similarity. The gray-scale depth map obtained from the point cloud, combining with the corresponding 2D texture, is used as the dual-channel input. The most important step of the preprocessing algorithm is the face hole filling, which provides a better intact face. The basic idea is to extract the 3D hole edge points, then project the hole edge points onto the 2D mesh plane to fill the hole points and map them back to the original 3D point cloud. Experiments were conducted to show the influence of depth map features and the size of training set on the accuracy of recognition rate.

\emph{Zhang Z, Da F, Yu Y.} \cite{zhang2019data} proposed a network structure similar to PointNet++. It contains three set abstraction (SA) modules: curvature-aware point sampler, neighbors grouper and multi-layer perceptrons (MLP). The first two SA are used to extract local features and the last one is used to aggregate global feature. The most important part of this work is the training set. All the training sets are unreal data, that are synthesized by sampling from a statistical 3D Morphable Model of face shape and expression based on the GPMM \cite{luthi2017gaussian}. This method addresses the problem of lacking a large training dataset. After classification training, triplet loss is used to fine-tune the network with real faces, which can get better results.

 \emph{Mu at al.} \cite{mu2019led3d} proposed a lightweight CNN for 3D face recognition, especially for low-quality data. This network contains 4 blocks which have 32, 64, 128, and 256 convolution filters, respectively. The feature maps from these four convolutional blocks are captured by different Receptive Fields (RFs), down sampled to a fixed size by max-pooling, and integrated to form another conversion block. This process is completed by the Multi-Scale Feature Fusion (MSFF) module. The aim is to efficiently improve the representation of low-quality face data. The Spatial Attention Vectorization (SAV) module is used to replace the Global Average Pooling (GAP) layer (also used by ResNet) to vectorize feature maps. The SAV highlights important spatial facial clues and conveys more discriminative cues by adding an attention weight map to each feature map \cite{mu2019led3d}. In addition, three methods are used to augment the training data: pose generating (by adjusting virtual camera parameters), shape jittering (by adding Gaussian noise to simulate rough surface changes), and shape scaling (by zooming in the depth face image with 1.1 times). At data preprocessing stage, similar to the above methods, a $10 \times 10$ patch surface is first cropped around a given nose tip with outliers removal. Then the cropped 3D point cloud is projected into a 2D space (depth surface) to generate a normal map image.
 
  \emph {Bhople at al.} \cite{bhoplepoint} proposed a network based on the PointNet architecture. It directly uses point cloud as input and uses the Siamese network for similarity learning. In addition, a way to augment database at the point cloud level is provided.
 
   \emph{Cao at al.} \cite{cao2020reliable} believed that the key to a reliable face recognition system is rich data sources. They used ANN as the network architecture, but paid more attention to data acquisition. A holoscopic 3D (H3D) face image database was created, which contains 154 raw H3D images. H3D imaging is recorded by using a regularly arranged array of small lenses, which are closely packed together and connected with a recording device \cite{cao2020reliable}. Therefore, it can display 3D images with continuous parallax and full-color images can be viewed in a wider viewing area. Wavelet transform is used for feature extraction, as it performs well in the presence of illumination changes and face orientation changes, also reduces image information redundancy and retains the most important facial features \cite{cao2020reliable}. This is definitely a new direction for 3D face recognition, but the accuracy of this method is quite low, only reaching slightly higher than 80\%.
   
   \emph{Dutta at al.} \cite{dutta2020sppcanet} proposed a lightweight sparse principal component analysis network (SpPCANet). It includes three parts: convolutional layer, nonlinear processing layer and feature merging layer. For data preprocessing, common ways are used to detect and crop the face area. First an ICP-based registration technology is used to register a 3D point cloud, then the 3D point cloud is converted into a depth image and finally all faces are cropped into rectangles based on the position of the nose tip. The system obtained 98.54\% recognition rate on Bosphorus3D.
   
\begin{table*}[ht]
    \centering
   \caption{Deep learning-based techniques.  }\label{table:deeplearning_methods}
    \begin{tabularx}{\textwidth}{>{\tr}p{0.12\textwidth} >{\tr}p{0.13\textwidth} >{\tr}p{0.15\textwidth} >{\tr}p{0.12\textwidth} >{\tr}p{0.14\textwidth} >{\tr}p{0.08\textwidth} >{\tc}p{0.11\textwidth}}
        \toprule
        Author/year & Network & Layers & Input & Matching & Database & RR (\%)(Rank-1) \\
        \midrule

        Kim et al. (2017) \cite{kim2017deep} & Finetuning VGG & 16 convolutional, 3 FC, 1 softmax & $224 \times 224 \times 3$ & Cosine distance & Bosphorus & 99.2 \\  \tableLine

        Zulqarnain et al. (2018) \cite{zulqarnain2018learning} & FR3DNet & 13 convolutional, 3 FC, 1 softmax & $160 \times 160 \times 3$ & Cosine distance & Texas-3D & 100.0 \\  \tableLine

        Feng et al. (2019) \cite{feng20193d} & ANN & 2 DCNN & Depth map & - & CASIA & 98.44 \\  \tableLine 

        Cai et al. (2019) \cite{cai2019fast} & Pre-ResNet-34, Pre-ResNet-24, Pre-ResNet-14 & - & $96 \times 96 \times 3$ & Euclidean distance & FRGC v2 & 100 \\  \tableLine

        Lin at al. (2019) \cite{lin2019local} & ResNet-18 & 17 convolutional, 1 FC & $256 \times 256 \times 3$ & Similarity tensor calculated from 2 feature tensors & Bosphorus & 99.71 \\  \tableLine

        Olivetti at al. (2019) \cite{olivetti2019deep} & MobileNetV2 & - & $224 \times 224 \times 3$ & - & Bosphorus & 97.56 \\  \tableLine
        
        Tan at al. (2019) \cite{tan2019face} & ResNet-18 & 17 convolutional, 1 FC & $256 \times 256 \times 3$ & Cosine distance & CASIA & 99.7 \\  \tableLine   
        
        Xu at al. (2019) \cite{xu20193d} & LeNet5 & Two parallel CNNs (4 convolutions, 4 pooling, and 1 FC) & RGB image with depth map & Euclidean distance & CASIA & -  \\  \tableLine 
        
        Zhang at al. (2019) \cite{zhang2019data} & Data-Free Point Cloud Net (similar to PointNet++) & 3 set abstraction (SA) modules & 3D Point Cloud & Cosine similarity & FRGC v2 & 98.73 \\  \tableLine 

         Mu at al. (2019) \cite{mu2019led3d} & - & 4 convolution blocks, a MSFF module and a SAV module & Low quality input & Cosine distance & Lock3DFace & 81.02 \\  \tableLine 

        Cao at al. (2020) \cite{cao2020reliable} & ANN & - & H3D image & - & H3D \cite{cao2020reliable} & - \\  \tableLine
         
        Dutta at al. (2020) \cite{dutta2020sppcanet} & SpPCANet & Convolution layer, nonlinear processing layer, feature-pooling layer & Depth images & linear SVM \cite{cortes1995support} & Frav3D & 96.93 \\
        
        \bottomrule
    \end{tabularx}
\end{table*}


\subsubsection*{Summary}
In this section, we review the deep learning-based 2D and 3D face recognition techniques. Most of the DCNN-based 3D methods achieve very high recognition accuracy and run at fast speed. For example, \cite{cai2019fast} only required 0.84 seconds to identify a target face from a gallery of 466 faces. \cite{zulqarnain2018learning} got 100\% recognition rate on Texas-3D. There are three important parts in a DCNN-based system: training set, data preprocessing and network architecture. The deep learning-based methods always require a large number of datasets to train the network. The lack of large-scale 3D face datasets is still an open problem in DCNN-based 3D face recognition research. Before passing the data to the network, appropriate data preprocessing can improve accuracy, as CNN usually has less tolerance for pose changes. In addition, adopting a suitable DCNN is also important. In the above reviewed works, most of them use a single CNN but few use dual CNN such as \cite{xu20193d}. The reorganization of CNNs may also be a topic of future research.

 

\section{Discussion}
\label{sec:Discussion}

In the past decade, 3D face recognition has achieved significant growth in 3D face databases, recognition rates, and robustness to face data variance, such as low-resolution, expression, pose and occlusion. The conventional methods and deep learning-based methods are thoroughly reviewed in Section 3 and Section 4, respectively. Based on the feature extraction algorithms, the conventional methods are divided into three types: local feature-based, holistic-based and hybrid methods.

\begin{itemize}
\item Local feature descriptors extract features from small regions of a 3D facial surface. In some cases, the region can be reduced to small patches around the detected key-points. Compared to the global descriptors for the holistic-based methods, the number of extracted local descriptors is related to the content of input face (entire or partial). It is commonly assumed that only a small number of facial regions are affected by occlusion, partial missing or distortion caused by data corruption, while most other regions persist unchanged. A face representation is derived from the combination of many local descriptors. Therefore, local facial descriptors are not compromised when dealing with the deformation of a few parts caused by facial expressions or occlusions \cite{lei2016two}.

\item A global representation is extracted from an entire 3D face, which usually makes the holistic-based methods compact and therefore computationally efficient. In addition, these methods can achieve great accuracy in the presence of complete neutral faces. However, they rely on the availability of full face scans and are sensitive to face alignment, occlusion, and data corruption. Therefore, face registration is a very important step for holistic-based methods.


\item The hybrid methods combine the algorithms of extracting local features and global features. They can handle more conditions, such as the pose variance and occlusion variance.
\end{itemize}

\begin{table*}[t]
    \centering
   \caption{RR (\%)(Rank-1) of DCNN-based methods on other databases. HQ: high quality image. LQ: low quality image. }\label{table:deeplearning_performance}
    \begin{tabularx}{\textwidth}{>{\tr}p{0.15\textwidth} >{\tc}p{0.06\textwidth} >{\tc}p{0.06\textwidth} >{\tc}p{0.06\textwidth} >{\tc}p{0.06\textwidth} >{\tc}p{0.06\textwidth} >{\tc}p{0.06\textwidth} >{\tc}p{0.06\textwidth} >{\tc}p{0.06\textwidth} >{\tc}p{0.06\textwidth} >{\tc}p{0.06\textwidth}}
        \toprule
        Reference & FRGC v2 & BU3D-FE & BU4D-FE & Bosphorus & CASIA & GavabDB & Texas-3D & 3D-TEC & UMBDB & ND-2006 \\
        \midrule

        Kim et al. \cite{kim2017deep} & - & 95.0 & - & 99.2 & - & - & - & 94.8 & - & - \\ \tableLine
                 
        FR3DNet\cite{zulqarnain2018learning} & 97.06 & 98.64 & 95.53 & 96.18 & 98.37 & 96.39 & 100.00 & 97.90 & 91.17 & 95.62 \\ \tableLine

        FR3DNetFT. \cite{zulqarnain2018learning} & 99.88 & 99.96 & 98.04 & 100.0 & 99.74 & 99.70 & 100.0 & 99.12 & 97.20 & 99.13 \\ \tableLine

        Feng et al. \cite{feng20193d} & - & - & - & - & 85.93 & - & - & - & - & - \\ \tableLine

        Cai et al. \cite{cai2019fast} & 100 & 99.88 & - & 99.75 & - & - & - & 99.07 & - & - \\ \tableLine

        Lin at al. \cite{lin2019local} & - & 96.2 & - & 99.71 & - & - & - & - & - & - \\ \tableLine
        
        Olivetti at al. \cite{olivetti2019deep} & - & - & - & 97.56 & - & - & - & - & - & - \\ \tableLine
        
        Tan at al. \cite{tan2019face} & - & - & - & 99.2 & 99.7 & - & - & - & 99.2 & - \\ \tableLine
        
        
        Zhang at al.\cite{zhang2019data} & 92.74 & - & - & 93.38& - & - & - & - & - & - \\ \tableLine
        Zhang at al. (ft)\cite{zhang2019data} & 98.73 & - & - & 97.5 & - & - & - & - & - & - \\ \tableLine
                
        Mu at al. \cite{mu2019led3d} & - & - & - & 91.27 (HQ)\newline 90.70 (LQ) & - & - & - & - & - & - \\ \tableLine
        
        Dutta at al. (2020) \cite{dutta2020sppcanet} & - & - & - & 98.54 & 88.80 & - & - & - & - & - \\
        


        \bottomrule
    \end{tabularx}
\end{table*}

Since 2016, research on DCNN-based 3D face recognition has been carried out. Table \ref{table:deeplearning_performance} summarizes the recognition rate of our surveyed methods on different databases under rank-1. Compared with the conventional face recognition algorithms, DCNN-based methods have the advantages of simpler pipelines and higher performance. In general, the deep learning-based methods do not have to perform keypoint detection, face segmentation or feature fusions. Instead, they only need to convert 3D data into a suitable network input format (e.g. 2D images). Moreover, since the pre-trained networks are often fine-tuned using the training data generated from 3D faces, the chance of delivering promising performance has been greatly improved. However, they are more reliant on the training set than the conventional methods. Therefore, data augmentation is one of the key challenges we are facing. Besides of the network structure design, data preprocessing also has a huge influence on the recognition performance.

To improve the accuracy and performance of face recognition systems, we discuss the following (future) directions by considering new face data generation, data preprocessing and DCNN design.

\begin{itemize}
\item Augmenting new face data. In Section 4, almost every proposed method provides a strategy for augmenting new face data. It is because a large amount of training data is required to train networks. A network trained with sufficient data can better distinguish features, while a small number of samples may cause overfitting. To synthesize new face data, one way is to use 3D deformable facial Model to generate new shape and expression (such as \cite{kim2017deep}, \cite{zhang2019data}). Another method is to randomly select sub-feature sets from different samples of a person and combine them to generate a new face. Some other methods augment the pose variance by rotating the original data (e.g. \cite{olivetti2019deep}).

\item Data Preprocessing. It is also a key point of improving face recognition accuracy. A well-known problem of the rigid ICP registration is that it cannot guarantee an optimal convergence \cite{kim2017deep}. This means it may not be possible to accurately register all 3D faces with different poses to the reference face. Furthermore, CNNs may not have much tolerance to pose deviation. Better conversion techniques (e.g. from 3D faces to 2D images) would also improve face recognition performance.

\item Applying appropriate loss functions. There are many CNNs available for 3D face recognition, such as VGG, ResNet, mobileNet and PointNet++. Most researchers adopt one of them as their network. Usually, the network is restructured by changing the FC layer and adding softmax. Recently, applying loss functions to supervise the network layers also has become one active research topic. Using effective loss functions can reduce the complexity of training and improve feature learning capabilities. For example, \cite{zhang2019data} adopted multiple loss functions to improve the extraction efficiency.

\item Creating large-scale 3D face database. Current 3D face databases are often smaller compared to the counterparts in 2D color face recognition, and nearly all the deep learning-based 3D face recognition methods fine-tune the pre-trained networks on the converted data from 3D faces. Hence, large-scale 3D face databases could enable training from scratch and improve the recognition difficulty, making it more closer to real-world applications.
\end{itemize}


Besides of the above aspects, researchers can consider combining conventional methods with CNN. For example, the keypoint detection techniques in conventional 3D face recognition methods can be incorporated into the deep learning-based methods to better attend area of interest. 3D face recognition methods for low-quality data (such as low resolution) also need more attention.

\section{Conclusion}
\label{sec:conclusion}

3D face recognition has become an active and popular research topic in the field of image processing and computer vision in recent years. In this paper, a summary of public 3D face databases is first provided, followed by a comprehensive survey on the 3D face recognition methods proposed in the past decade. The recognition methods are divided into two categories based on their feature extraction methods: conventional and deep learning-based. The conventional techniques are further classified into local-based, holistic-based and hybrid methods. We reviewed these methods by comparing their performance on different databases, the computational cost, and the robust against the expressions, occlusions and pose variations. From the above literature review, we found that local methods can better handle face expressions and occluded images at the cost of higher computation comparing with holistic-based methods. Hybrid methods combined the local and global features can achieve better performance and address the challenges including pose variations, illumination changes and facial expressions, etc. 

We reviewed recent advances in 3D face recognition based on deep learning, mainly focusing on face augmentation, data preprocessing and network architectures. CNN is one of the most popular deep neural networks for 3D face recognition. According to the network adopted, the deep learning-based 3D face recognition methods are broadly divided into VGG-based, ResNet-based and others. With these powerful networks, the performance of 3D face recognition has been greatly improved.

We also discussed the involved characteristics and challenges, and provided potential future directions for 3D face recognition. For instance, large-scale 3D face databases are in great need to advance 3D face recognition in the future. We believe our survey will provide comprehensive information to readers and inspire insights in the community.

\bibliographystyle{IEEEtran}
\bibliography{reference.bib}

\begin{biography}[{\includegraphics[width=1in,height=1.2in,clip,keepaspectratio]{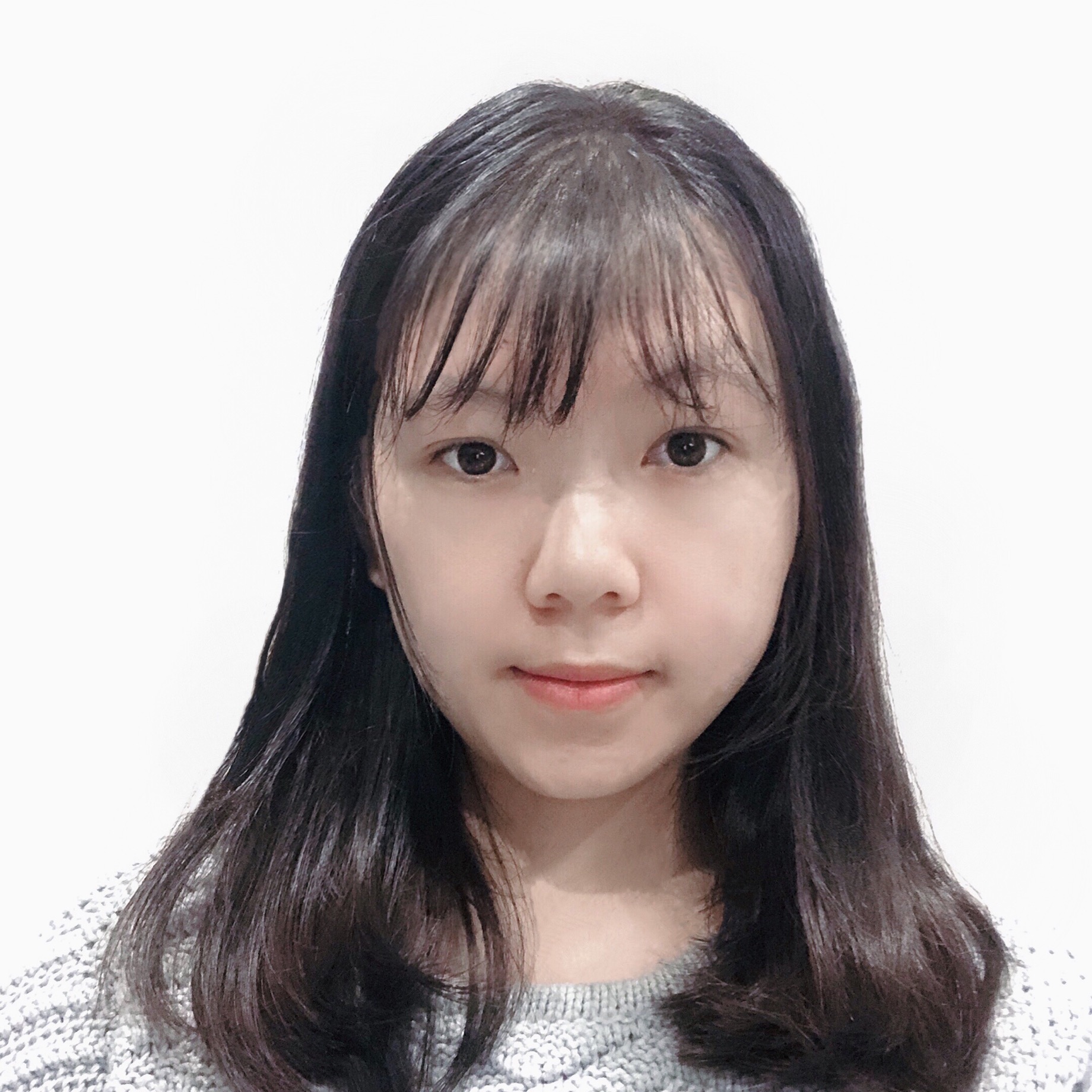}}]{Yaping Jing} received the Bachelor degree (Hons) of Information Technology from Deakin University, Australia in 2016. She is currently a PhD candidate in School of Information Technology at Deakin University. Her research interests include 3D face recognition, 3D data processing and machine learning.
\end{biography}

\begin{biography}[{\includegraphics[width=1in,height=1.2in,clip,keepaspectratio]{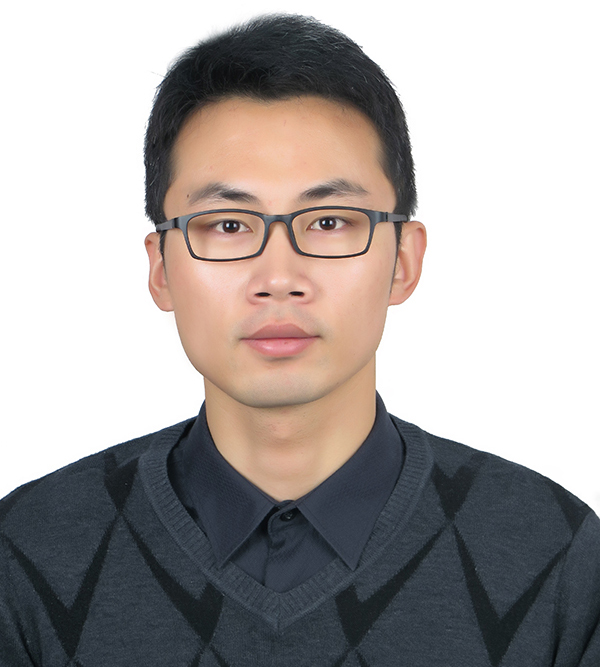}}]{Xuequan Lu} is a Lecturer (Assistant Professor) at Deakin University, Australia. He spent more than two years working as a Research Fellow in Singapore. Prior to that, he received his Ph.D from Zhejiang University (China) in June 2016. His research interests mainly fall into the category of visual data computing, for example, geometry modeling, processing and analysis, animation/simulation, 2D data processing and analysis. More information can be found at \url{http://www.xuequanlu.com}.
\end{biography}

\begin{biography}[{\includegraphics[width=1in,height=1.2in,clip,keepaspectratio]{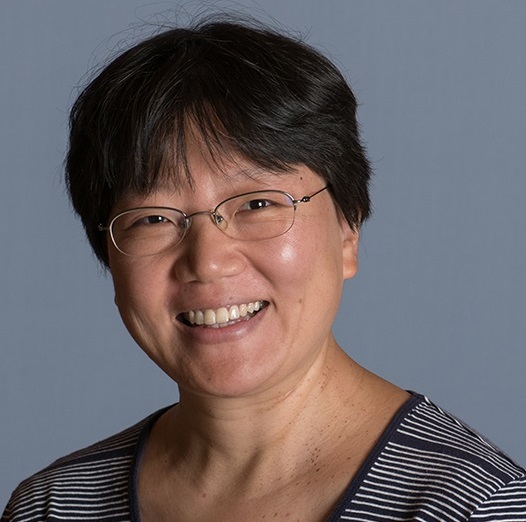}}]{Shang Gao} received her Ph.D. degree in computer science from Northeastern University, China in 2000. She is currently a senior Lecture in School of Information Technology, Deakin University, Australia. Her current research interests include cybersecurity, cloud computing and machine learning.
\end{biography}

\end{document}